\documentclass[11pt]{article}

\usepackage[final]{acl}

\usepackage{times}
\usepackage{latexsym}

\usepackage[T1]{fontenc}

\usepackage[utf8]{inputenc}

\usepackage{microtype}

\usepackage{inconsolata}

\usepackage{tabularx}
\usepackage{makecell}
\usepackage{graphicx}

\usepackage{cuted}
\usepackage{caption}
\usepackage[table]{xcolor}
\usepackage{xcolor}
\usepackage{colortbl}   
\usepackage{array,booktabs}
\usepackage{makecell}   
\usepackage{pifont}     
\usepackage{adjustbox}  
\usepackage{array} 
\usepackage{bbm}
\usepackage{amsmath}
\usepackage{amssymb}
\usepackage{multirow}

\usepackage{listings}
\usepackage{xcolor}

\lstdefinestyle{promptstyle}{
  basicstyle=\ttfamily\scriptsize,
  breaklines=true,
  breakatwhitespace=true,
  frame=single,
  columns=fullflexible,
  keepspaces=true,
  xleftmargin=1mm,
  xrightmargin=1mm
}

\definecolor{Hdr}{gray}{0.95}
\definecolor{RowHL}{gray}{0.90}
\definecolor{SCbg}{HTML}{FFF5CC}
\definecolor{SChead}{HTML}{FFE082}

\renewcommand{\checkmark}{\textcolor{green!60!black}{\ding{51}}}
\newcommand{\xmark}{\textcolor{red!70!black}{\ding{55}}}

\newcolumntype{S}{>{\columncolor{SCbg}}c} 
\renewcommand{\arraystretch}{1.15}
\title{WaymoQA: A Multi-View Visual Question Answering Dataset for Safety-Critical Reasoning in Autonomous Driving}

\author{
  \vspace{2mm}
  \textbf{Seungjun Yu\textsuperscript{1}},
  \textbf{Seonho Lee\textsuperscript{2}},
  \textbf{Namho Kim\textsuperscript{3}},
  \textbf{Jaeyo Shin\textsuperscript{1}},
\\[-1mm]
  \textbf{Junsung Park\textsuperscript{1}},
  \textbf{Wonjeong Ryu\textsuperscript{1}},
  \textbf{Raehyuk Jung\textsuperscript{1}},
  \textbf{Hyunjung Shim\textsuperscript{1}\thanks{Corresponding Author.}}
\\[1mm]
  \textsuperscript{1}KAIST,
  \textsuperscript{2}KRAFTON,
  \textsuperscript{3}Hanyang University,
  Seoul, Republic of Korea
\\[-1mm]
  \small{
  \texttt{
  seungjunyu@kaist.ac.kr,\quad
  glanceyes@krafton.com,\quad
  knhnh@hanyang.ac.kr,\quad
  jaeyo\_shin@kaist.ac.kr
  }
  }
\\[-1mm]
  \small{
  \texttt{
  jshackist@kaist.ac.kr,\quad
  petac@kaist.ac.kr,\quad
  mulnyangi@naver.com,\quad
  kateshim@kaist.ac.kr
  }
  }
}

\begin{document}
\maketitle
\vspace{-12mm}

\begin{strip}
\centering
\vspace{-4.0mm}
\includegraphics[width=1.0\textwidth]{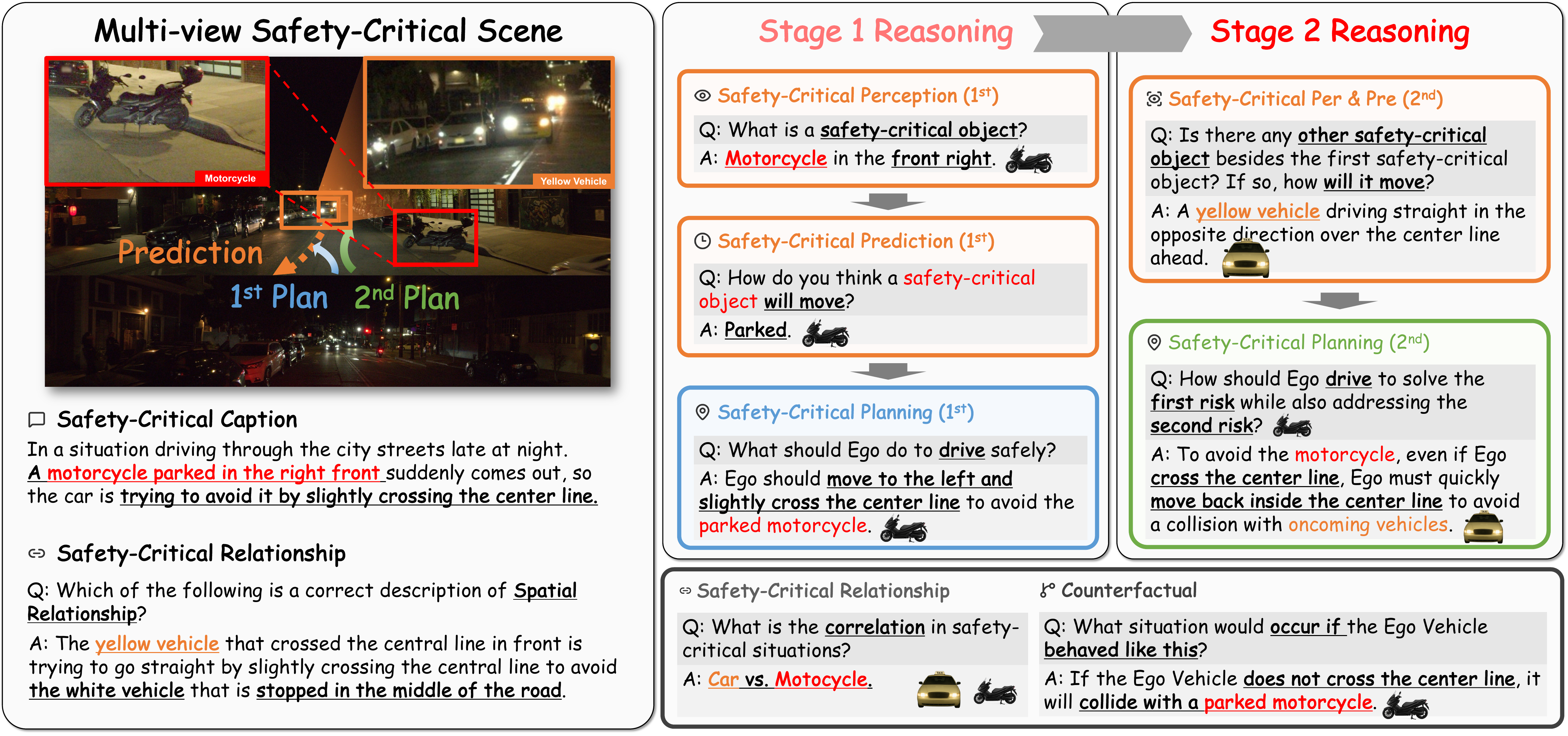}
\captionof{figure}{WaymoQA overview of action-induced two-stage safety-critical reasoning.
A Stage~1 maneuver resolves the primary hazard but induces a secondary risk, requiring Stage~2 planning under the changed risk landscape.}
\label{fig:teaser}
\vspace{-4.0mm}
\end{strip}
\begin{abstract}
In real-world driving, resolving a primary hazard often precipitates a secondary one: swerving to avoid a stalled vehicle may force the ego car into oncoming traffic. Existing driving QA datasets fail to capture this cascading causality, evaluating perception, prediction, and planning in silos without anticipating the risks action induces. We introduce WaymoQA, a multi-view dataset that frames safety-critical reasoning as a two-stage task: resolving the immediate threat, then mitigating the action-induced secondary risk. WaymoQA provides 35,000 human-annotated QA pairs across image and video modalities, with 28K training and 6.4K test examples. Experiments show that SOTA models struggle on safety-critical scenes, while fine-tuning on WaymoQA improves both reasoning accuracy and consistency. WaymoQA thus serves as a rigorous benchmark and resource for safety-aware autonomous driving. Our code and data are provided in \url{https://github.com/sjyu001/WaymoQA}.
\end{abstract}
\section{Introduction}
\label{sec:intro}

Multimodal Large Language Models (MLLMs) have rapidly advanced from general vision-language tasks to autonomous driving, they now support end-to-end planning, scenario generation, and vision-language action modeling~\cite{feng2025verdi,fu2024drive,guo2025vdt,huang2024making,ma2024dolphins,shao2024lmdrive}. 
Within this landscape, Driving Question Answering (Driving QA) has emerged as a promising framework for evaluating high-level scene understanding and decision-making~\cite{marcu2024lingoqa, li2025fine}, since QA format directly exposes what a model perceives, predicts, and decides.

Driving QA datasets have evolved from single-view setups such as DRAMA~\cite{malla2023drama} and DriveLM~\cite{sima2024drivelm} to multi-camera frameworks like NuScenes-QA~\cite{qian2024nuscenes} and NuPlan-QA~\cite{park2025nuplanqa}, progressively improving spatial coverage and inter-agent modeling.
However, both generations share a common blind spot: they are overwhelmingly built from normal driving scenarios~\cite{fu2025video,li2024mvbench,liu2024mmbench,zeng2025vision}, leaving safety-critical events, the rare but consequential moments that most challenge real-world deployment, largely untested.

\begin{figure}
  \centering
  \includegraphics[width=\linewidth]{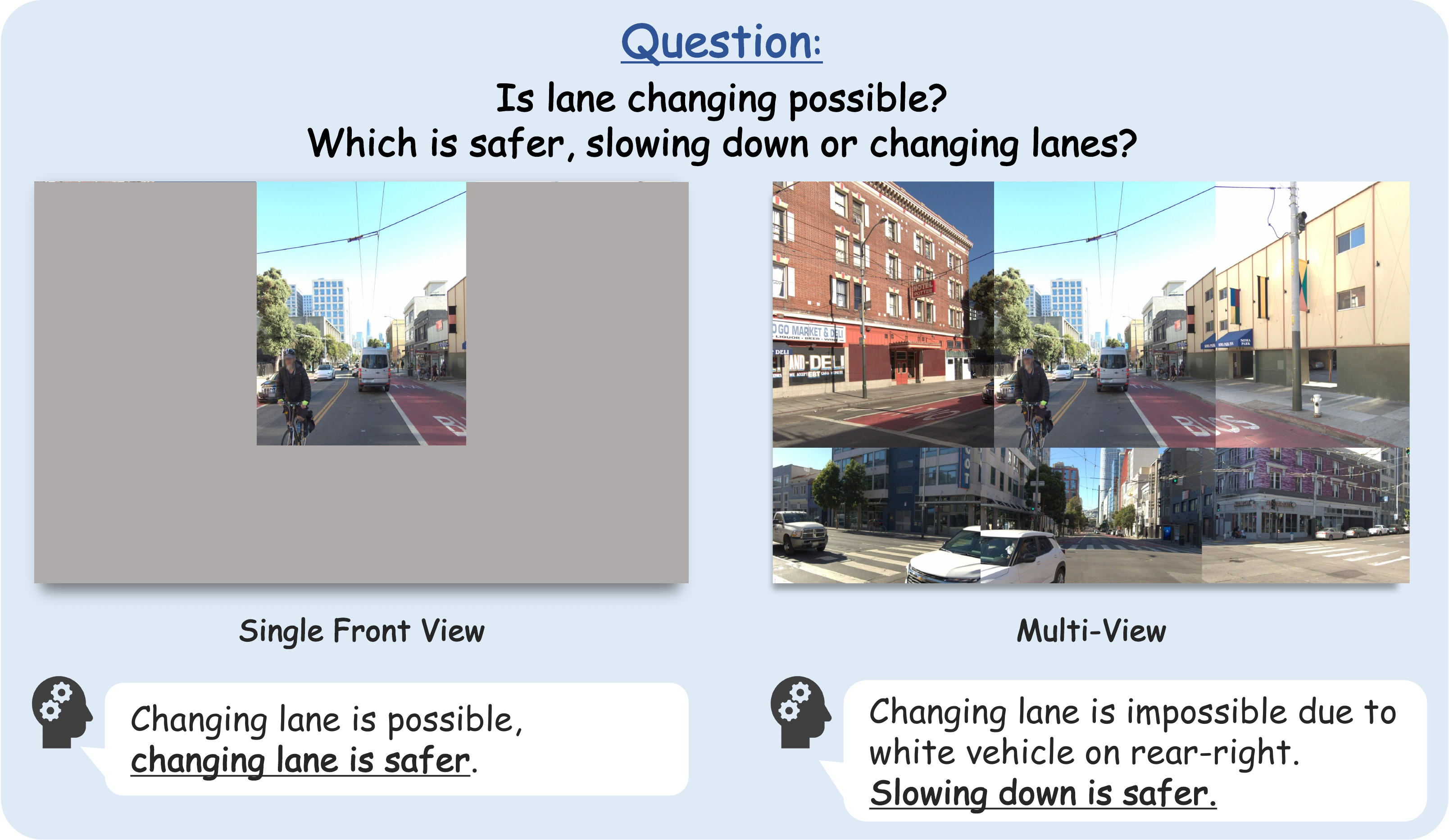}
  \vspace{-5mm}
  \caption{\textbf{Same-Scene View Comparison.}}
  \label{fig:compare_vew}
  \vspace{-5mm}
\end{figure}

Recent benchmarks have begun addressing this gap. 
NAVSAFE~\cite{sima2025centaur}, DVBench~\cite{zeng2025vision}, and VRU-Accident~\cite{kim2025vru} introduce safety-critical evaluation for driving VQA, yet three key limitations persist. 
(\textbf{L1}) They are mainly evaluation-only and do not provide training data for learning safety-critical reasoning. 
(\textbf{L2}) They often rely on limited or single-view inputs, which can miss blind spots and occluded agents; and 
(\textbf{L3}) They treat perception, prediction, and planning as independent questions, without evaluating action-induced downstream risks where resolving one hazard can create another.
As shown in Fig.~\ref{fig:compare_vew}, the same scene can lead to opposite decisions depending on view coverage: a lane change appears feasible from the front view, but multi-view input reveals a rear-side vehicle that would turn that maneuver into a collision. 
This illustrates that safety-critical reasoning requires both multi-view evidence and explicit reasoning over how a chosen action reshapes the surrounding risk landscape.

To address these limitations, we introduce WaymoQA, the first training-enabled, safety-critical multi-view driving QA dataset. 
WaymoQA is built from rare pre-crash scenarios in the Waymo End-to-End (Waymo E2E) dataset~\citep{xu2025wod} and filtered
using the 47 U.S. NHTSA pre-crash scenario types \citep{national2007pre}.
Here, \emph{long-tail} specifically refers to rare pre-crash scenario types defined by the NHTSA taxonomy, rather than rare
environmental conditions such as adverse weather, nighttime driving, or heavy traffic.
WaymoQA contains 35,000 human-written question-answer pairs across safety-critical videos and manually selected key frames. 
It provides 28,585 training examples and 6,415 test examples, supporting both open-ended QA and multiple-choice formats. Each scene is observed through eight synchronized camera views covering front, side, and rear directions, supplying the surrounding context that single-view setups miss.

Central to WaymoQA is its formulation of \emph{Safety-Critical Reasoning} as a compositional two-stage task. 
Stage 1 resolves the immediate risk by identifying the safety-critical object, predicting its behavior, and selecting a feasible safe action. 
Stage 2 then addresses the secondary risks induced by the Stage 1 action, reflecting real-world driving where mitigating one hazard can introduce another. 
For example, as shown in Fig.~\ref{fig:teaser}, the ego vehicle first avoids a parked motorcycle by briefly crossing the center line, but must then return to its lane before colliding with an oncoming vehicle. 
WaymoQA further combines two complementary QA modalities: Video QA captures realized temporal events, while Image QA supports decision-oriented and counterfactual reasoning from key frames, broadening the space of possible actions and outcomes beyond the single realized trajectory.

Comprehensive experiments show that pretrained MLLMs consistently struggle with safety-critical reasoning, underperforming markedly on safety-critical questions compared to normal scenes. 
Fine-tuning on WaymoQA improves ImageQA, VideoQA, open-ended QA, and two-stage consistency, indicating gains beyond answer-option selection toward free-form and decision-conditioned reasoning. 
Transfer, ablation, and robustness analyses further show that WaymoQA supervision improves external driving QA benchmarks and planning metrics, benefits from multi-view inputs, and remains stable under answer-order permutation and environmental variations.

Our contributions are threefold: 
(1) we define multi-view safety-critical reasoning for autonomous-driving VQA; 
(2) we introduce WaymoQA, the first training-enabled safety-critical multi-view driving QA dataset with image/video and open-ended/MCQ formats; and 
(3) we present MCQ, consistency, transfer, ablation, and robustness evaluations showing that WaymoQA supervision improves safety-critical reasoning.
\begin{figure*}
  \centering
  \includegraphics[width=0.95\linewidth]{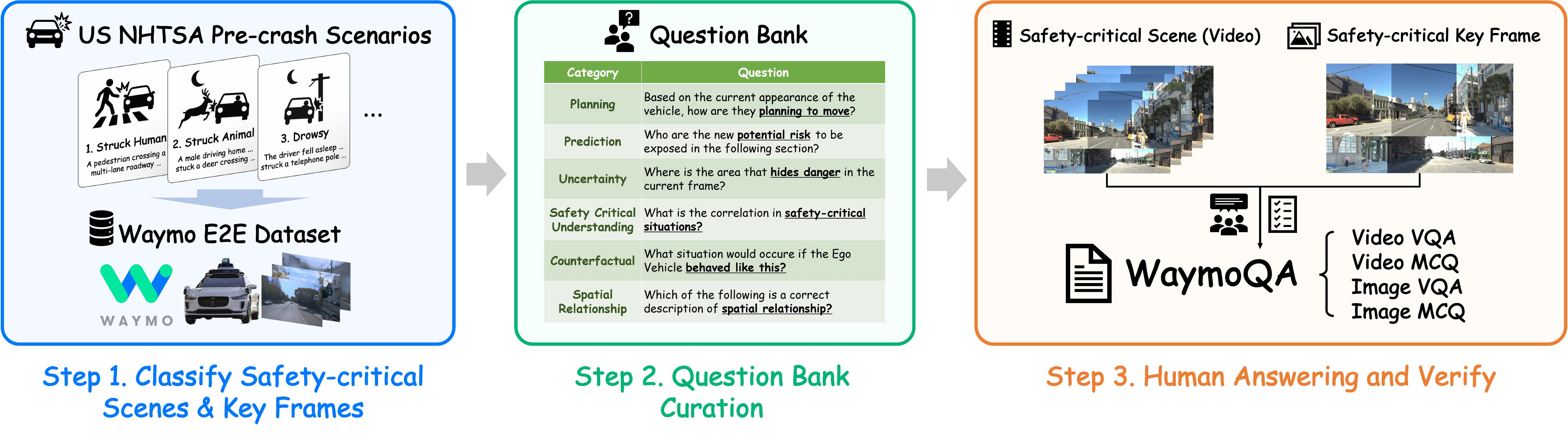}
  \vspace{-3mm}
  \caption{\textbf{WaymoQA construction pipeline.} We filter safety-critical Waymo E2E scenes, curate a structured QA bank, and perform human answering and verification to produce Video/Image VQA and MCQ splits.}
  \vspace{-3mm}
  \label{fig:main}
\end{figure*}

\section{Related Works}
\label{sec:relatedworks}

\textbf{MLLM in Autonomous Driving.}
MLLMs have shown strong high-level reasoning across diverse tasks~\cite{liu2024chain,stone2023open,chen2024spatialvlm,hong2024cogagent}, motivating their use in autonomous driving~\cite{chen2024driving,cui2024drive,zhou2025autovla}. 
Recent studies extend MLLMs from perception and captioning to driving-oriented reasoning, planning, and trajectory generation, including spatial/action reasoning~\cite{tian2024drivevlm,xu2024drivegpt4}, high-resolution or 3D-enhanced perception~\cite{ding2025hilm,han2025dme}, generative planning~\cite{hu2023gaia,mao2023gpt}, agentic driving~\cite{jin2024surrealdriver}, VLA models~\cite{zhou2025opendrivevla}, and text-to-trajectory conversion~\cite{hwang2024emma}. 
Overall, the field is shifting from scene narration toward decision-making coupled with perception and control.

\textbf{Safety-Critical Autonomous Driving.}
Safety-critical driving is difficult due to rarity, diversity, and causal dependencies. 
Prior work has studied planner-driven stress testing~\cite{rempe2022generating,zhong2022guided,zhong2023language}, adversarial, controllable, and RL-based scenario generation~\cite{xieadvdiffuser,lin2025causal,liu2024safety}, real-video anomaly analysis~\cite{yao2022dota}, safety-focused datasets and benchmarks~\cite{luo2023simulation,wang2024deepaccident}, and simulator-scale synthesis~\cite{zhang2024chatscene,dosovitskiy2017carla}. 
Recent efforts further improve realism and multi-view alignment through diffusion, reinforcement learning, and causal rationales~\cite{li2025avd2,zheng2024open,zhou2025safemvdrive,shao2024deepseekmath,fang2024abductive}.
\section{WaymoQA Dataset}
\label{sec:dataset}

We describe WaymoQA, a human-annotated multi-view VQA dataset for safety-critical reasoning. 
Figure~\ref{fig:main} summarizes the construction pipeline.

\begin{table*}[t]
\centering
\renewcommand{\arraystretch}{0.95} 
\resizebox{\textwidth}{!}{%
\begin{tabular}{l | *{6}{c} | *{6}{c} | c}
\toprule
\multicolumn{1}{c|}{\textbf{Dataset}} &
\textbf{Normal} &
\makecell{\textbf{Safety-}\\\textbf{Critical}} &
\makecell{\textbf{Multi}\\\textbf{View}} &
\makecell{\textbf{Train}\\\textbf{Data}} &
\textbf{Video} &
\textbf{Image} &
\textbf{Per.} &
\makecell{\textbf{Nor.}\\\textbf{Rea.}} &
\makecell{\textbf{SC.}\\\textbf{Rea.}} &
\makecell{\textbf{2SS.}\\\textbf{Rea.}} &
\makecell{\textbf{Uncer-}\\\textbf{tanty}} &
\makecell{\textbf{Counter-}\\\textbf{factual}} &
\textbf{Amount} \\
\midrule
NuScenes-QA~\cite{qian2024nuscenes}   & \checkmark  & \xmark & \checkmark & \checkmark   & \xmark & \checkmark & \checkmark & \xmark & \xmark & \xmark & \xmark & \xmark & 460K \\
NuplanQA~\cite{park2025nuplanqa}         & \checkmark  & \xmark & \checkmark & \checkmark   & \xmark & \checkmark & \checkmark & \xmark & \xmark & \xmark & \xmark & \xmark & 1M \\
DriveLM~\cite{sima2024drivelm}                & \checkmark   & \xmark & \checkmark & \xmark  & \xmark & \checkmark & \checkmark & \checkmark & \xmark & \xmark & \xmark & \xmark & 443K \\
DriveMLLM~\cite{guo2024drivemllm}                        & \checkmark  & \xmark & \xmark & \xmark  & \checkmark & \xmark & \checkmark & \xmark & \xmark & \xmark & \xmark & \xmark & 4K \\
BDD-X~\cite{kim2018textual}                       & \checkmark  & \xmark  & \xmark & \checkmark & \checkmark & \checkmark & \checkmark & \xmark & \xmark & \xmark & \xmark & \xmark & 26K \\
DRAMA~\cite{malla2023drama}                           & \checkmark & \xmark  & \xmark & \checkmark & \checkmark & \checkmark & \checkmark & \xmark & \xmark & \xmark & \xmark & \xmark & 14K \\
DriveBench~\cite{xie2025vlms}  & \checkmark & \xmark  & \checkmark & \xmark & \xmark & \checkmark & \checkmark & \checkmark & \xmark & \xmark & \xmark & \xmark & 20K \\
VRU-Accident~\cite{kim2025vru}   & \checkmark  & \checkmark & \xmark & \xmark   & \checkmark & \checkmark & \checkmark & \checkmark & \xmark & \xmark & \xmark & \xmark & 6K \\
DVBench~\cite{zeng2025vision}   & \checkmark  & \checkmark & \xmark & \xmark   & \checkmark & \checkmark & \checkmark & \checkmark & \xmark & \xmark & \xmark & \xmark & 10K \\
STRIDE-QA~\cite{ishihara2026stride} & \checkmark & \xmark & \checkmark & \checkmark 
& \checkmark & \checkmark 
& \checkmark & \checkmark & \xmark 
& \xmark & \xmark & \xmark & 16M \\
\midrule
\rowcolor{RowHL}
\textbf{WaymoQA}      & \textbf{\checkmark} & \textbf{\checkmark} & \textbf{\checkmark} & \textbf{\checkmark} &
\textbf{\checkmark} & \textbf{\checkmark} & \textbf{\checkmark} &
\textbf{\checkmark} & \textbf{\checkmark} & \textbf{\checkmark} & \textbf{\checkmark} & \textbf{\checkmark} & \textbf{35K}\\
\bottomrule
\end{tabular}}
\vspace{-3mm}
\caption{\textbf{Comparison of driving QA datasets and reasoning coverage.} Abbreviations: Perception(Per.), Normal Reasoning(Nor.Rea.), Safety-Critical Reasoning(SC.Rea.), Two-Stage Safety-Critical Reasoning(2SS.Rea.).}
\label{tab:dataset_comp}
\vspace{-3mm}
\end{table*}

\subsection{Scene Selection and Multi-View Inputs}

We build WaymoQA from the Waymo End-to-End (Waymo E2E) dataset~\cite{xu2025wod}, which contains long-tail real-world driving scenarios. 
To reduce subjectivity, we select candidate safety-critical scenes using the 47 pre-crash scenario types defined by the U.S. National Highway Traffic Safety Administration (NHTSA)~\cite{national2007pre}, providing reproducible criteria for collision-relevant risks, abrupt agent interactions, and hazardous traffic configurations.
Final safety-critical/normal classification and key-frame selection are performed by human annotators, without the use of an automatic model.

Each scene has synchronized multi-view observations: Image QA uses eight same-timestamp views, and Video QA uses multi-view clips from the same scenario. 
This coverage is essential because action feasibility may depend on agents beyond the front view; for example, a lane change may appear safe from the front camera but become unsafe once a rear-side vehicle is visible.

Safety-critical events are brief, often lasting about 2--4 seconds in a 22-second Waymo E2E scenario. 
Sampling critical moments would overrepresent hazardous states and weaken normal-to-critical evaluation. 
We therefore curate key frames covering both contexts while avoiding near-duplicates, using a \(2{:}3\) normal-to-critical ratio, one-second minimum interval, and at least five key frames per video. 
This preserves temporal context for Video QA while yielding balanced image-level questions.

To prevent data leakage, we split the dataset at the Waymo scenario-token level, such that all clips, key frames, and QA pairs derived from the same scenario are assigned to a single split.
The train, validation, and test splits contain 127, 148, and 54 unique scenarios, respectively, with zero scenario-token overlap across the three splits.

\subsection{Question Bank Curation}

Before authoring QA pairs, we construct a \textbf{62-item question bank} to standardize coverage across input formats, scene types, and safety-critical reasoning categories. 
The bank follows a three-level hierarchy of input format, scene type, and reasoning category, covering Caption, Scene Understanding, Perception, Prediction, Planning, Uncertainty, Counterfactual, Special Event, Spatial Relationship, Temporal Grounding, and Traffic Signal. 
Each template is labeled with structured metadata, with details provided in Appendix~\ref{sup:details}.

Using this curated question bank, WaymoQA contains 35K human-authored QA pairs, split into 28,585 training and 6,415 test examples.
The training split contains open-ended QA pairs that provide rich supervision for model fine-tuning.
The test split contains human-written four-option multiple-choice questions (MCQs), enabling objective evaluation with a fixed answer key.
Table~\ref{tab:dataset_comp} compares WaymoQA with prior driving QA datasets.
Unlike previous datasets, WaymoQA jointly provides safety-critical scenarios, multi-view inputs, trainable annotations, image/video QA, counterfactual questions, and two-stage safety-critical reasoning.


\begin{figure*}[!t]
  \centering
  \includegraphics[width=1.0\linewidth]{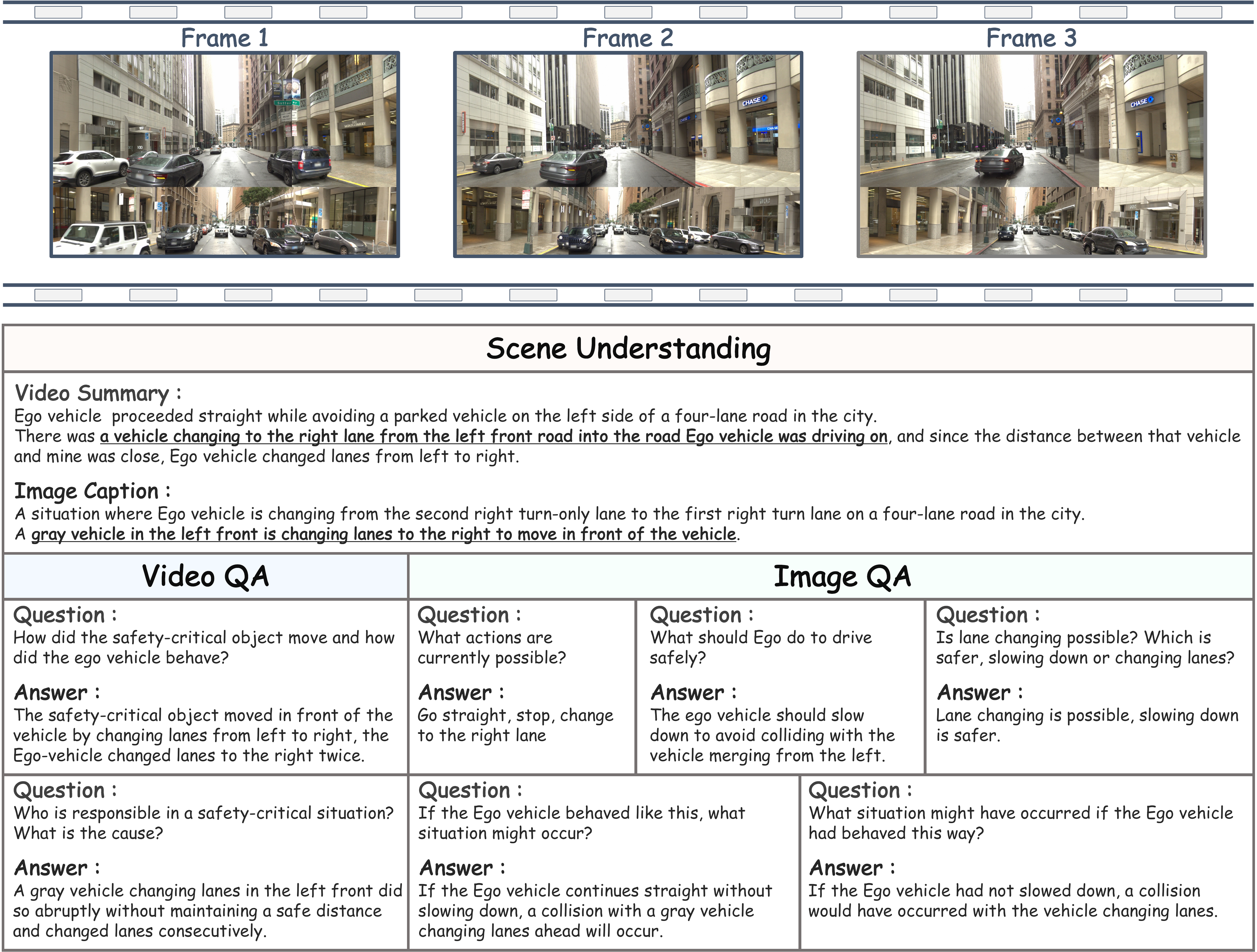}
  \vspace{-3mm}
  \caption{Examples of Video/Image QA in WaymoQA.}
  \label{fig:video_image}
  \vspace{-3mm}
\end{figure*}




\subsection{Two-Stage Safety-Critical Reasoning}
\label{sec:two-stage}

WaymoQA formulates safety-critical driving QA as a two-stage action-induced reasoning task.
Stage~1 asks the model to identify the primary hazard, predict its behavior, and select a feasible action to resolve the immediate risk.
Stage~2 is induced by this Stage~1 action, as the initial maneuver may change the ego vehicle's position and interaction with surrounding agents, exposing secondary risks that were not the primary threat before.
The model must then choose an action that mitigates the downstream risk while preserving the Stage~1 objective.

Unlike ordinary single-step scene understanding, the Stage~2 answer cannot be inferred from the current scene alone; it depends causally on the Stage~1 maneuver.
For example, crossing the center line may be necessary to avoid a parked motorcycle, but this maneuver can induce a new conflict with an oncoming vehicle.
Stage~2 evaluates whether the model can anticipate how its action changes the future risk landscape, rather than merely recognizing visible objects or predicting the next frame.

This design separates immediate-risk errors from downstream-foresight errors and supports counterfactual branching, where different Stage~1 actions induce different Stage~2 risks.
WaymoQA evaluates not only hazard identification, but also causal reasoning over action-induced safety consequences.

\subsection{Video vs. Image QA}
Safety-critical scenarios are crucial but rare in practice, making it difficult to achieve broad coverage when constructing reasoning data from video alone.
We therefore leverage two complementary perspectives.
Video QA enables temporal reasoning about what actually occurred, while Image QA extends safety-critical reasoning by exploring alternative decisions from a single key frame.

In the example of Fig.~\ref{fig:video_image}, a vehicle cuts in from the left as the ego changes lanes, and the ego moves to the right lane to avoid a collision. 
From a single key frame of the same situation, Image QA can ask which actions are feasible at this time, which option is safer given visible risks, and whether a lane change is possible or slowdown is preferable. 
Since these questions move beyond the single realized path, they expand the set of alternative actions and near term outcomes for the same scene.

Combining the two perspectives increases the diversity of reasoning signals for rare long-tail safety-critical events. This enables more robust learning and evaluation even when scenario data are limited.

\subsection{Formats and Evaluation}
WaymoQA provides two complementary formats. Open-ended QA pairs are used for training, whereas four-option multiple-choice questions (MCQs) are used for validation and testing.
Importantly, models are fine-tuned exclusively on open-ended QA pairs and are never exposed to MCQ answer options or distractors during training. Therefore, improvements in MCQ evaluation cannot result from direct memorization of WaymoQA-specific distractor
patterns during fine-tuning.
We adopt a four-option MCQ design and write all options by humans. 
The position of the correct answer is uniformly distributed across the four slots to avoid position bias and reduce random guessing. 
This choice avoids LLM-as-a-judge scoring and enables objective accuracy evaluation with a fixed answer key.
Prior work~\cite{chen2024mllm,chen2504llm,wataoka2024self} has shown that LLM-based grading can exhibit biases, such as a preference for longer answers regardless of correctness and a tendency to favor responses written in a style similar to the evaluator. 
Our MCQ setup is therefore intended to provide a consistent and fair evaluation without relying on an LLM grader.

\begin{table}[!t]
\centering
\small
\setlength{\tabcolsep}{4pt}
\resizebox{\columnwidth}{!}{%
\begin{tabular}{l|c}
\toprule
Statistic & Value \\
\midrule
Annotators & 7 \\
Independent annotations per scene & \(\geq 3\) \\
Annotation manager flagging rate & 26\% \\
Re-annotation threshold & Fleiss' \(\kappa < 0.7\) \\
ImageQA \(\kappa\) Normal / Others / Safety-Critical & 0.95 / 0.90 / 0.97 \\
VideoQA \(\kappa\) Normal / Others / Safety-Critical & 0.90 / 0.91 / 0.98 \\
Independent verifiers & 3 \\
Verification flag rate & 34\% \(\rightarrow\) 3\% \(\rightarrow\) 0\% \\
\bottomrule
\end{tabular}%
}
\vspace{-3mm}
\caption{Annotation reliability and verification statistics.}
\label{tab:annotation_quality}
\vspace{-5mm}
\end{table}

\begin{table*}[!t]
\centering
\setlength{\tabcolsep}{2.2pt}
\renewcommand{\arraystretch}{1.0}
\resizebox{\textwidth}{!}{%
\begin{tabular}{l | c | c | *{7}{S} | c || c | c | *{6}{S} | c}
\toprule
\multicolumn{1}{l|}{\textbf{Model}}
& \multicolumn{10}{c||}{\textbf{ImageQA}}
& \multicolumn{9}{c|}{\textbf{VideoQA}} \\
\cmidrule(lr){2-11}\cmidrule(lr){12-20}
& \multicolumn{1}{c|}{\textbf{Normal}}
& \multicolumn{1}{c|}{\textbf{Others}}
& \multicolumn{7}{>{\columncolor{SChead}}c|}{\textbf{Safety-Critical}}
& \multicolumn{1}{c||}{\textbf{Overall}}
& \multicolumn{1}{c|}{\textbf{Normal}}
& \multicolumn{1}{c|}{\textbf{Others}}
& \multicolumn{6}{>{\columncolor{SChead}}c|}{\textbf{Safety-Critical}}
& \multicolumn{1}{c}{\textbf{Overall}} \\
\cmidrule(lr){4-10}\cmidrule(lr){14-19}
& Overall & Overall
& SU & PE & PR & PL & SC & UN & Overall
& 
& Overall & Overall
& SC & SU & PE & TG & UN & Overall
& \\
\midrule
Human Experts 
& 96.1 & 96.3 & 96.8 & 98.0 & 96.4 & 94.3 & 98.7 & 94.9 & 96.0 & 96.1
& 96.9 & 96.6 & 97.2 & 97.2 & 96.0 & 94.0 & 92.5 & 95.9 & 96.5 \\

GPT 5.2
& 75.7 & 57.0 & 68.4 & 65.6 & 73.8 & 62.3 & 57.6 & 53.0 & 64.5 & 67.4
& 79.1 & 78.8 & 54.1 & 65.5 & 50.0 & 49.0 & 81.3 & 61.6 & 72.6 \\
\midrule

LLaVA1.5(7B)
& 35.9 & 37.5 & 24.5 & 21.3 & 24.7 & 25.5 & 27.2 & 17.0 & 23.2 & 30.7
& 42.4 & 34.7 & 13.5 & 27.5 & 32.6 & 27.4 & 33.3 & 27.5 & 35.9 \\

LLaVA-OneVision(7B)
& 66.7 & 51.7 & 50.8 & 42.2 & 56.9 & 53.2 & 45.5 & \textbf{64.9} & 54.4 & 58.6
& 65.8 & 82.1 & 32.4 & 43.4 & 40.3 & 29.4 & 62.5 & 42.3 & 59.2 \\

InternVL3.5(8B)
& 40.1 & 46.6 & 26.6 & 30.4 & 23.7 & 41.1 & 25.3 & 22.2 & 28.9 & 37.0
& 45.1 & 50.8 & 29.7 & 28.2 & 51.9 & 19.6 & 43.7 & 32.9 & 41.4 \\

InternVL3.5(4B)
& 38.0 & 42.3 & 21.7 & 24.1 & 23.6 & 39.2 & 41.1 & 20.2 & 27.8 & 34.4
& 36.3 & 58.4 & 37.8 & 23.4 & 42.3 & 17.6 & 39.5 & 29.3 & 36.7 \\

InternVL3(8B)
& 33.6 & 47.0 & 28.8 & 24.9 & 27.1 & 46.2 & 41.7 & 27.8 & 32.9 & 35.9
& 36.3 & 39.7 & 40.5 & 27.5 & 50.0 & 25.4 & 56.2 & 36.2 & 36.8 \\

InternVL3(14B)
& 40.2 & 55.6 & 23.4 & 32.8 & 52.1 & 45.3 & 45.8 & 39.0 & 42.1 & 44.0
& 46.6 & 60.4 & \underline{59.4} & 27.3 & 51.9 & 25.0 & 57.4 & 38.6 & 46.3 \\

Qwen2.5-VL(32B)
& 71.2 & 60.3 & 52.6 & 44.6 & \underline{74.3} & \underline{64.6} & 56.4 & 61.4 & 62.6 & 65.4
& 72.3 & 79.7 & 43.2 & 46.2 & 51.9 & 31.3 & 67.8 & 47.6 & 64.1 \\

Qwen2.5-VL(7B)
& 68.1 & 55.6 & 54.4 & 55.3 & 71.0 & 64.0 & 43.0 & 54.6 & 61.1 & 62.4
& \underline{72.7} & 64.4 & 32.4 & 53.1 & 55.7 & 31.3 & 64.5 & 49.5 & 63.8 \\

Phi3.5-vision
& 48.9 & 44.5 & 32.0 & 25.6 & 41.8 & 45.6 & 44.3 & 46.5 & 40.9 & 44.7
& 50.9 & 52.5 & 37.8 & 31.7 & 32.6 & 24.5 & 54.1 & 32.6 & 45.1 \\

Gemma3(4B)
& 59.7 & 48.7 & 34.2 & 37.1 & 53.8 & 44.3 & 51.3 & 33.8 & 43.5 & 50.6
& 60.0 & 66.1 & 54.0 & 29.6 & 38.5 & \underline{54.9} & 54.2 & 41.1 & 53.8 \\

Gemma3(27B)
& 64.3 & 62.6 & \underline{67.2} & 61.6 & 67.1 & 59.4 & 54.4 & 49.4 & 60.4 & 62.3
& 65.1 & 69.4 & 43.2 & 53.7 & 55.7 & 39.2 & 62.5 & 51.8 & 60.7 \\
\midrule

InternVL2-DA-DriveLM(4B)
& 54.2 & 37.7 & 44.4 & 24.1 & 41.2 & 26.8 & 38.6 & 46.1 & 36.8 & 43.6
& 57.0 & 52.5 & 37.8 & 43.4 & 30.7 & 27.4 & 58.3 & 40.5 & 50.3 \\

OpenDriveVLA
& 54.9 & 43.9 & 57.1 & 50.7 & 68.3 & 53.1 & 41.1 & 49.6 & 56.0 & 53.4
& 67.9 & 72.0 & 43.2 & 49.7 & 50.0 & 35.3 & 54.1 & 47.5 & 60.9 \\
\midrule

LLaVA-OneVision(7B) (FT, Ours)
& \underline{74.3} & 62.8 &  64.7 & \underline{61.9} & 67.4 & 61.9 & \underline{59.3} & 59.3 & \underline{65.5} & \underline{68.3} 
& 71.1 & \textbf{84.5} & 54.2 & \underline{60.7} & 59.2 & 44.1 & \underline{68.1} & \underline{58.3} & \underline{68.1} \\

InternVL3.5(8B) (FT, Ours)
& 61.6 & \underline{65.6} & 57.1 & 54.1 & 53.2 & 50.3 & 51.3 & 54.2 & 53.1 & 58.7 
& 62.1 & 76.1 & 51.3 & 51.3 & \underline{60.1} & 37.2 & 63.7 & 52.4 & 60.4 \\

Qwen2.5-VL(7B) (FT, Ours)
& \textbf{74.9} & \textbf{72.7} & \textbf{76.1} & \textbf{76.6} & \textbf{77.8} & \textbf{65.9} & \textbf{67.8} & \underline{63.8} & \textbf{71.2} & \textbf{72.9}
& \textbf{76.0} & \underline{83.1} & \textbf{63.5} & \textbf{66.2} & \textbf{63.5} & \textbf{58.9} & \textbf{70.8} & \textbf{65.0} & \textbf{72.8} \\

\bottomrule
\end{tabular}%
}
\vspace{-3mm}
\caption{Detailed ImageQA and VideoQA performance of MLLMs on WaymoQA. 
\textbf{Bold}: best; \underline{underline}: second best. 
Abbreviations: SC (Scene Caption), SU (Scene Understanding), 
PE (Perception), PR (Prediction), PL (Planning), TG (Temporal Grounding), UN (Uncertainty), FT (Fine-tuning).}
\label{tab:image_video_qa}
\vspace{-3mm}
\end{table*}

\subsection{Human Answering and Verification}

Recent studies have begun to automatically generate driving QA data with MLLMs~\cite{khalili2025autodrive,parikh2025roadsocial}. 
However, such data may reinforce pretrained model biases, reduce diversity, and contribute to model collapse~\cite{shumailov2024ai}. 
We therefore employ seven Ph.D. annotators with prior knowledge of driving and autonomous-driving research for internal human annotation and verification.
Annotation was conducted internally as part of the research project, with no separate per-item payment.

As summarized in Table~\ref{tab:annotation_quality}, each scene is annotated by at least three annotators, with one annotation manager flagging guideline violations. 
Scenes with Fleiss' \(\kappa < 0.7\) or ambiguous answers are re-annotated, resulting in agreement above 0.90 across all scene types in both Image QA and Video QA.
Three independent verifiers further check ambiguity, excessive difficulty, and multiple plausible answers, reducing the verification flag rate from 34\% to 0\% after revision.


\section{Experiments}

\subsection{Experimental Settings}
We assess multiple families and sizes of state-of-the-art MLLMs and expert models, including variants of LLaVA~\cite{liu2024improved,li2024llava}, InternVL~\cite{wang2025internvl3,zhu2025internvl3,gao2024mini}, Qwen-VL~\cite{bai2025qwen2}, Phi3.5-vision~\cite{abdin2024phi3technicalreporthighly}, Gemma3~\cite{team2025gemma}, and OpenDriveVLA~\cite{zhou2025opendrivevla}.

For \textbf{Image QA}, each example consists of eight synchronized camera views covering the front, side, and rear directions. 
For \textbf{Video QA}, each video clip spans 14 seconds, we uniformly sample 50 frames per video for evaluation.

\subsection{Results of Image QA and Video QA}

Table~\ref{tab:image_video_qa} reports the main benchmark results. 
Across pretrained models, safety-critical performance is consistently lower than normal-scene performance, showing that safety-critical reasoning requires more than generic scene recognition: models must interpret agent behavior, feasible actions, and downstream consequences.

Among pretrained models, Qwen2.5-VL performs best in both modalities. 
Qwen2.5-VL(32B) achieves the best pretrained overall accuracy, followed by Qwen2.5-VL(7B). 
However, most pretrained models remain weak on safety-critical questions, especially in VideoQA, suggesting that temporal context alone does not ensure reliable risk-aware reasoning.

Fine-tuning on WaymoQA substantially improves performance. 
Qwen2.5-VL(7B) improves from 62.4\% to 72.9\% on ImageQA and from 63.8\% to 72.8\% on VideoQA, with large gains on safety-critical questions. 
It also surpasses GPT-5.2 under the same MCQ protocol, suggesting that targeted WaymoQA supervision can be more effective than general-purpose model scale alone. 
Additional fine-tuning results on LLaVA-OneVision(7B) and InternVL3.5(8B) show consistent gains, indicating that WaymoQA provides transferable supervision across MLLM families.

\subsection{Open-ended Evaluation}
\label{sec:open_ended_eval}

MCQ accuracy is our primary protocol because it provides objective fixed-answer evaluation without relying on LLM-as-a-judge scoring. 
This is especially important for safety-critical questions, where open-ended automatic metrics can obscure differences among weak pretrained models. 
We therefore use open-ended evaluation only as a complementary analysis of free-form reasoning.

As shown in Table~\ref{tab:open_ended_eval}, WaymoQA fine-tuning substantially improves GPT-based scoring, BLEU, ROUGE, and BERTScore, achieving the best overall and Safety-Critical scores across most metrics. 
However, pretrained-model trends are less clear than in the MCQ benchmark, supporting our use of human-written MCQ as the main evaluation protocol. 
These results indicate that WaymoQA supervision improves not only answer selection but also free-form safety-critical reasoning.
\begin{table}[!h]
\centering
\setlength{\tabcolsep}{2.0pt}
\renewcommand{\arraystretch}{1.0}
\resizebox{\columnwidth}{!}{%
\begin{tabular}{l|cccc|cccc}
\toprule
\multirow{2}{*}{\textbf{Model}}
& \multicolumn{4}{c|}{\textbf{Overall}}
& \multicolumn{4}{c}{\textbf{Safety-Critical}} \\
\cmidrule(lr){2-5}\cmidrule(lr){6-9}
& GPT & B-4 & R-L & BERT
& GPT & B-4 & R-L & BERT \\
\midrule
LLaVA-OneVision(7B)      & 25.2 & 1.43 & 10.6 & 85.1 & 25.5 & 0.86 & 10.5 & 85.2 \\
InternVL3.5(8B)          & 25.3 & 1.59 & 11.5 & 84.6 & 20.0 & 1.32 & 11.4 & 84.8 \\
Qwen2.5-VL(7B)           & 25.5 & 1.60 & 11.3 & 84.7 & 23.0 & 1.12 & 10.2 & 84.7 \\
Qwen2.5-VL(32B)          & 25.1 & 0.85 & 10.2 & 83.7 & 19.9 & 0.09 & 8.8  & 82.7 \\
Gemma3(27B)              & 25.5 & 1.45 & 11.7 & 85.0 & 21.3 & 1.32 & 11.7 & 85.1 \\
InternVL2-DA-DriveLM(4B) & 23.1 & 2.96 & 15.7 & 85.9 & 17.5 & 2.70 & 15.8 & 85.8 \\
\midrule
Qwen2.5-VL(7B) (FT)      & \textbf{40.4} & \textbf{7.25} & \textbf{29.0} & \textbf{88.1} 
                         & \textbf{34.0} & \textbf{4.72} & \textbf{23.1} & \textbf{87.1} \\
\bottomrule
\end{tabular}%
}
\vspace{-3mm}
\caption{Open-ended evaluation on WaymoQA. Abbreviations: B-4(BLEU-4), R-L(ROUGE-L).}
\label{tab:open_ended_eval}
\vspace{-3mm}
\end{table}

\subsection{Two-stage Consistency Evaluation}
WaymoQA formulates safety-critical reasoning as a two-stage action-induced process, where the Stage~1 maneuver induces or exposes the Stage~2 risk (Sec.~\ref{sec:two-stage}).

We evaluate this dependency on a safety-critical ImageQA subset containing 245 key frames with paired Stage~1 and Stage~2 questions.
Both questions are anchored to the same key frame, identified by the same Waymo scenario token and frame index.
Each Stage~2 question is paired with its corresponding Stage~1 question from the same key frame, such that Stage~2 reasoning is conditioned on the Stage~1 maneuver.
The evaluation contains 394 paired instances, consisting of 244 prediction chains and 150 planning chains.
VideoQA is excluded from this evaluation because equivalent paired two-stage chains were not constructed for the video subset.

To verify whether models truly follow this dependency, we introduce a stricter \emph{joint correctness} evaluation.
Specifically, a two-stage instance is counted as correct only if \textbf{both} Stage~1 and Stage~2 answers match the ground truth:
\begin{equation}
\label{eq:two_stage_joint_acc}
\mathrm{Acc}_{\mathrm{2\text{-}stage}}
=
\frac{1}{N}\sum_{i=1}^{N}
\mathbbm{1}\Big[\hat{y}^{(1)}_{i}=y^{(1)}_{i}\ \wedge\ \hat{y}^{(2)}_{i}=y^{(2)}_{i}\Big].
\end{equation}
This joint criterion captures the causal dependency of sequential risks; Stage~2 is evaluated together with the corresponding Stage~1 decision rather than as an independent question.
As shown in Table~\ref{tab:two_stage_eval}, this joint evaluation causes a substantial drop for generic baselines (typically 7--14\%), indicating difficulty in maintaining stage-to-stage consistency when the dependency is enforced.
In contrast, the WaymoQA-fine-tuned Qwen2.5-VL-7B exhibits a markedly smaller degradation, suggesting that WaymoQA supervision improves \emph{decision-conditioned} reasoning across immediate risk resolution (Stage~1) and induced secondary risk mitigation (Stage~2).




\begin{table}[!h]
\centering

\small
\setlength{\tabcolsep}{4pt}
\renewcommand{\arraystretch}{1.0}
\begin{tabular}{lccc}
\toprule
\textbf{Model}
& \textbf{Single-stage}
& \textbf{Two-stage}
& $\Delta$ \\
\midrule
GPT-5.2                    & 67.8 & 60.9 & \textcolor{red}{-6.9} \\
LLaVA-OneVision-7B         & 53.0 & 38.9 & \textcolor{red}{-14.1} \\
Qwen2.5-VL-32B             & 65.5 & 54.6 & \textcolor{red}{-10.9} \\
Gemma3-27B                 & 63.1 & 49.5 & \textcolor{red}{-13.6} \\
OpenDriveVLA               & 59.3 & 52.1 & \textcolor{red}{-7.2} \\
\midrule
\textbf{Qwen2.5-VL-7B (FT)}
                          & \textbf{72.8}
                          & \textbf{70.4}
                          & \textbf{\textcolor{red}{-2.4}} \\
\bottomrule
\end{tabular}
\vspace{-3mm}
\caption{Two-stage consistency evaluation.}
\label{tab:two_stage_eval}
\vspace{-3mm}
\end{table}

\subsection{Controlled Supervision Analysis}

An improvement after fine-tuning on WaymoQA could in principle arise simply from exposure to additional driving-domain QA data, rather than from safety-critical supervision itself. 
We therefore conduct a matched-control experiment using the same Qwen2.5-VL-7B model, LoRA configuration, training budget, and evaluation protocol, while varying only the training data.

As shown in Table~\ref{tab:control}, removing safety-critical examples from WaymoQA decreases accuracy by 4.6 points overall and 5.6 points on safety-critical questions. 
A size-matched NuScenes-QA control performs substantially worse, trailing full WaymoQA training by 11.9 and 12.9 points, respectively. 
These results indicate that the observed improvement cannot be explained solely by additional driving-domain supervision and that safety-critical examples provide a specific benefit for risk-focused reasoning.

\begin{table}[!h]
\centering
\small
\begin{tabular}{lcc}
\toprule
\textbf{Training Data}
& \textbf{Overall}
& \textbf{Safety-Critical} \\
\midrule
WaymoQA (Full)          & \textbf{67.8} & \textbf{65.8} \\
WaymoQA (Normal-only)   & 63.2 & 60.2 \\
NuScenes-QA (Matched)   & 55.9 & 52.9 \\
\midrule
$\Delta$ Full -- Normal-only
                        & +4.6 & +5.6 \\
$\Delta$ WaymoQA -- NuScenes-QA
                        & +11.9 & +12.9 \\
\bottomrule
\end{tabular}
\caption{Matched-control analysis using the same Qwen2.5-VL-7B model and LoRA training setup.}
\label{tab:control}
\end{table}

\subsection{Transfer and Generalization}
\label{sec:transfer_generalization}

We further examine whether WaymoQA supervision transfers beyond the main QA benchmark. 
First, we evaluate planning performance on the Waymo End-to-End validation split. 
As shown in Table~\ref{tab:transfer_generalization}, WaymoQA supervision improves ADE and RFS even without trajectory supervision, suggesting that safety-critical reasoning provides useful high-level decision signals for downstream driving behavior. 
Combining WaymoQA with trajectory supervision yields the best performance, indicating that language-based safety reasoning and geometric motion supervision are complementary.

Second, we evaluate cross-dataset transfer on DriveLM-nuScenes and NuPlanQA. 
As shown in Table~\ref{tab:transfer_generalization}, Qwen2.5-VL(7B) fine-tuned on WaymoQA consistently improves over the base model on external driving QA datasets. 
This suggests that the learned reasoning patterns are not narrowly overfitted to the WaymoQA distribution, but transfer to other autonomous-driving QA settings.

Finally, we conduct a preliminary closed-loop evaluation on Bench2Drive to examine whether WaymoQA supervision transfers beyond offline QA and open-loop planning.
Using the same trajectory-supervised policy, WaymoQA improves the Driving Score from 65.0 to 71.2 and the Success Rate from 8.3\% to 16.7\% on 12 randomly selected routes.
On 16 safety-critical routes, the Driving Score improves from 57.4 to 63.4 and the Success Rate from 6.2\% to 12.5\%.
These results provide preliminary evidence that WaymoQA supervision transfers to downstream closed-loop behavior; however, given the limited number of evaluated routes, we treat them as supporting evidence rather than proof of deployable driving safety.






\begin{table}[!h]
\centering
\setlength{\tabcolsep}{4pt}
\renewcommand{\arraystretch}{1.05}
\vspace{-2mm}

{\small
\begin{tabular}{cc|cc}
\toprule
\textbf{WaymoQA} & \textbf{Trajectory} 
& \textbf{ADE (5s) $\downarrow$} & \textbf{RFS $\uparrow$} \\
\midrule
\xmark & \xmark & 14.30 & 5.37 \\
\checkmark & \xmark & 11.60 & 5.91 \\
\xmark & \checkmark & 4.23 & 6.37 \\
\checkmark & \checkmark & \textbf{3.81} & \textbf{6.91} \\
\bottomrule
\end{tabular}
}

\vspace{0.5em}

\resizebox{\columnwidth}{!}{%
\begin{tabular}{lccc}
\toprule
\textbf{Model} 
& \textbf{DriveLM MCQ} 
& \textbf{DriveLM GPT} 
& \textbf{NuPlan MCQ} \\
\midrule
Qwen2.5-VL(7B)       & 47.1 & 16.0 & 41.8 \\
Qwen2.5-VL(7B) (FT)  & \textbf{58.8} & \textbf{26.7} & \textbf{54.8} \\
\bottomrule
\end{tabular}%
}

\vspace{0.5em}

{\small
\begin{tabular}{lcc}
\toprule
\textbf{Bench2Drive Setting}
& \textbf{Base+Traj}
& \textbf{+WaymoQA+Traj} \\
\midrule
Random (12 routes)
& 65.0 / 8.3
& \textbf{71.2 / 16.7} \\
Safety-critical (16 routes)
& 57.4 / 6.2
& \textbf{63.4 / 12.5} \\
\bottomrule
\end{tabular}
}

\vspace{-2mm}
\caption{Transfer and generalization beyond the main WaymoQA benchmark. Bench2Drive results report Driving Score / Success Rate (\%).}
\label{tab:transfer_generalization}
\vspace{-5mm}
\end{table}

\section{Analysis}
\subsection{Design Ablations}
\label{sec:design_ablation}
We analyze WaymoQA's core design choices, focusing on view coverage and temporal sampling.

\noindent\textbf{Multi-view input.}
We compare single front view, front-3 views, and full 8-view inputs. As shown in Table~\ref{tab:design_ablation}, performance consistently improves as view coverage increases, especially when moving from front-only views to full surrounding views. This supports our motivation that safety-critical reasoning often depends on side and rear evidence that is unavailable from the front camera alone.

\noindent\textbf{Temporal sampling rate.}
We compare our default 50-frame VideoQA setting, corresponding to approximately 1.2 FPS, with a denser 10 FPS setting. Table~\ref{tab:design_ablation} shows only marginal differences across representative models, suggesting that 1.2 FPS is sufficient for the high-level reasoning targeted by WaymoQA. We emphasize that WaymoQA is a reasoning benchmark, not a proxy for real-time low-level control.

\begin{table}[!h]
\centering
\setlength{\tabcolsep}{2.5pt}
\renewcommand{\arraystretch}{1.0}
\resizebox{\columnwidth}{!}{%
\begin{tabular}{l|ccc|cc}
\toprule
\multirow{2}{*}{\textbf{Model}}
& \multicolumn{3}{c|}{\textbf{View Ablation} (ImageQA/VideoQA)}
& \multicolumn{2}{c}{\textbf{FPS Ablation}} \\
\cmidrule(lr){2-4}\cmidrule(lr){5-6}
& Front & Front-3 & Full-8 & 1.2 FPS & 10 FPS \\
\midrule
LLaVA-OneVision(7B)      & 54.2 / 55.1 & 56.5 / 56.7 & 58.6 / 59.2 & 59.2 & 59.1 \\
InternVL3.5(8B)          & 35.7 / 37.2 & 36.8 / 38.9 & 37.0 / 41.4 & 41.4 & 42.1 \\
Qwen2.5-VL(7B)           & 58.1 / 56.2 & 58.8 / 59.3 & 62.4 / 63.8 & 63.8 & 64.0 \\
Qwen2.5-VL(32B)          & 62.3 / 61.4 & 63.7 / 62.4 & 65.4 / 64.1 & 64.1 & 64.4 \\
Gemma3(27B)              & 58.3 / 56.2 & 60.1 / 58.9 & 62.3 / 60.7 & 60.7 & 61.1 \\
Qwen2.5-VL(7B) (FT)      & \textbf{70.8 / 71.3} & \textbf{71.1 / 72.1} & \textbf{72.9 / 72.8} & \textbf{72.8} & \textbf{72.9} \\
\bottomrule
\end{tabular}%
}
\vspace{-3mm}
\caption{View and temporal sampling ablations.}
\label{tab:design_ablation}
\vspace{-3mm}
\end{table}

\subsection{Evaluation Robustness and Artifact Controls}
\label{sec:evaluation_robustness}

To test answer-position bias, we evaluate each question under 10 random option permutations. 
Table~\ref{tab:option_permutation} shows only small variance across models and modalities.

\noindent\textbf{Question-only baseline.}
Qwen2.5-VL-7B achieves 39.7\% without visual input and 40.5\% after WaymoQA fine-tuning, whereas multimodal accuracy improves from 62.4\% to 72.9\% on ImageQA and from 63.8\% to 72.8\% on VideoQA.
This suggests that the gains primarily rely on visual grounding rather than textual priors.

\noindent\textbf{Blind surface-cue analysis.}
Simple heuristics based on question-option overlap, alphabetical order, punctuation, digit count, or fixed option selection achieve only 23.4--26.0\%, close to the 25\% chance level.

\noindent\textbf{Distractor-swapped evaluation.}
On a held-out subset with regenerated distractors, the fine-tuning gain is preserved: overall accuracy improves from 62.7\% to 70.6\% with original distractors and from 75.0\% to 83.4\% with regenerated distractors.
Thus, the improvement is not specific to the original distractor construction.

Together with the open-ended evaluation in Sec.~\ref{sec:open_ended_eval}, these results support the reliability of our MCQ protocol across answer permutations, textual-prior controls, distractor variations, and free-form generation.

\begin{table}[!h]
\centering
\small
\setlength{\tabcolsep}{3.5pt}
\renewcommand{\arraystretch}{1.0}
\resizebox{\columnwidth}{!}{%
\begin{tabular}{lcc|cc}
\toprule
\textbf{Model}
& \multicolumn{2}{c|}{\textbf{ImageQA}}
& \multicolumn{2}{c}{\textbf{VideoQA}} \\
\cmidrule(lr){2-3}\cmidrule(lr){4-5}
& Mean & Var. & Mean & Var. \\
\midrule
LLaVA-OneVision(7B)      & 58.4 & 0.13 & 58.9 & 0.07 \\
InternVL3.5(8B)          & 37.0 & 0.02 & 41.1 & 0.03 \\
Qwen2.5-VL(7B)           & 62.4 & 0.09 & 63.4 & 0.04 \\
Qwen2.5-VL(32B)          & 65.2 & 0.06 & 64.1 & 0.02 \\
Gemma3(27B)              & 62.1 & 0.04 & 60.9 & 0.06 \\
InternVL2-DA-DriveLM(4B) & 43.4 & 0.11 & 50.1 & 0.02 \\
Qwen2.5-VL(7B) (FT)      & \textbf{72.9} & 0.09 & \textbf{72.7} & 0.03 \\
\bottomrule
\end{tabular}%
}
\vspace{-3mm}
\caption{Robustness to answer-option ordering. We report mean and variance over 10 random permutations.}
\label{tab:option_permutation}
\vspace{-3mm}
\end{table}

\subsection{Environmental and Statistical Reliability}
\label{sec:environment_reliability}
We analyze whether WaymoQA exhibits environmental imbalance and whether model performance varies across conditions. The dataset naturally reflects common real-world conditions, with more examples from bright daytime, sunny or overcast weather, light traffic, high visibility, and dry roads. Rarer conditions such as heavy rain, dark night, heavy traffic, and low visibility are less represented.

We further evaluate model performance across time of day, weather, road type, traffic density, visibility, and road condition. As shown in Table~\ref{tab:environment_breakdown}, performance varies across conditions, confirming that environment-aware analysis is important. At the same time, the relative ranking of models is broadly stable, and the WaymoQA fine-tuned model performs best across most conditions.

\begin{table}[!h]
\centering
\small
\setlength{\tabcolsep}{3pt}
\renewcommand{\arraystretch}{0.9}
\resizebox{\columnwidth}{!}{%
\begin{tabular}{ll|ll|ll}
\toprule
\textbf{Attribute} & \textbf{QA \%} &
\textbf{Attribute} & \textbf{QA \%} &
\textbf{Attribute} & \textbf{QA \%} \\
\midrule
Bright day & 57.6 & Sunny & 52.7 & Urban residential & 38.2 \\
Overcast day & 20.0 & Overcast & 27.4 & Intersection & 30.3 \\
Lit night & 17.1 & Partly cloudy & 7.8 & Urban downtown & 17.9 \\
Dusk/dawn & 3.9 & Light rain & 5.6 & Suburban & 8.5 \\
Dark night & 1.5 & Clear & 3.7 & & \\
\midrule
Light traffic & 78.4 & High visibility & 86.2 & Dry road & 90.4 \\
Moderate traffic & 17.4 & Medium visibility & 13.4 & Wet road & 9.6 \\
Empty traffic & 4.0 & Low visibility & 0.4 & & \\
Heavy traffic & 0.1 & & & & \\
\bottomrule
\end{tabular}%
}
\vspace{-3mm}
\caption{Environment distribution of WaymoQA.}
\label{tab:environment_breakdown}
\vspace{-3mm}
\end{table}
To further test robustness, we conduct an augmentation-based evaluation on a 1K test-QA subset by generating environmental variants of the same scenes, including dusk/dawn, foggy, nighttime, overcast, rainy, and snowy conditions. Table~\ref{tab:env_augmentation} shows that performance degrades under adverse conditions, with different perturbations affecting Normal Scene and Safety-Critical questions differently. This suggests that environmental robustness remains an important future direction.

Finally, because some fine-grained sub-categories contain relatively few samples, we conduct repeated evaluations on small sub-categories and report the mean and variance in Table~\ref{tab:small_category_reliability}. The observed variances are small, suggesting that the main performance gaps are stable rather than artifacts of random fluctuations from small sample sizes.

\begin{table}[!h]
\centering
\tiny
\setlength{\tabcolsep}{1.8pt}
\renewcommand{\arraystretch}{0.86}
\resizebox{\columnwidth}{!}{%
\begin{tabular}{lcccccc}
\toprule
\textbf{Model}
& \textbf{D/D}
& \textbf{Fog}
& \textbf{Night}
& \textbf{Over}
& \textbf{Rain}
& \textbf{Snow} \\
\midrule
LLaVA-OV
& -3.6/-2.4 & -6.6/-1.7 & -3.6/-2.0 & -1.8/-6.5 & -4.1/-6.8 & -1.8/-7.5 \\
Qwen-7B
& -5.7/-2.7 & -7.7/-0.0 & -4.1/-3.5 & -4.1/-3.5 & -5.2/-5.9 & -3.1/-4.6 \\
Qwen-32B
& -8.2/-0.3 & -9.1/-2.4 & -4.1/-0.0 & -4.3/-0.6 & -4.3/-7.3 & -3.4/-5.9 \\
Gemma3
& -7.1/-4.0 & -8.7/-3.1 & -3.9/-1.3 & -4.1/-2.7 & -3.9/-5.1 & -3.7/-2.4 \\
Qwen-FT
& -5.5/-7.5 & -6.7/-7.8 & -4.3/-5.6 & -9.2/-6.2 & -4.0/-3.9 & -5.9/-7.3 \\
\midrule
Avg.
& -6.02/-3.38 & -7.76/-3.00 & -4.00/-2.48 & -4.70/-3.90 & -4.30/-5.80 & -3.58/-5.54 \\
\bottomrule
\end{tabular}%
}
\vspace{-3mm}
\caption{Accuracy drop under environment augmentation on a 1K test-QA subset. Each cell reports $\Delta$ Normal / $\Delta$ Safety-Critical. Abbreviations: D/D (Dusk/Dawn), Fog (Foggy), Over (Overcast).}
\label{tab:env_augmentation}
\vspace{-3mm}
\end{table}
\begin{table}[!h]
\centering
\small
\setlength{\tabcolsep}{2.2pt}
\renewcommand{\arraystretch}{0.95}
\resizebox{\columnwidth}{!}{%
\begin{tabular}{llcccc}
\toprule
\textbf{Cat.} & \textbf{Sub.}
& \textbf{LLaVA1.5}
& \textbf{InternVL3}
& \textbf{InternVL3.5}
& \textbf{Qwen FT} \\
\midrule
Safety-Critical & SC & 13.8 / 0.6 & 40.5 / 0.0 & 29.7 / 0.0 & 63.5 / 0.0 \\
Safety-Critical & PE & 32.6 / 0.0 & 50.0 / 0.0 & 52.1 / 0.31 & 63.5 / 0.0 \\
Safety-Critical & TG & 27.4 / 0.0 & 25.4 / 0.0 & 19.6 / 0.0 & 58.9 / 0.0 \\
Safety-Critical & UN & 33.3 / 0.0 & 56.2 / 0.0 & 43.7 / 0.0 & 71.0 / 0.39 \\
\bottomrule
\end{tabular}%
}
\vspace{-3mm}
\caption{Repeated-run reliability on small VideoQA sub-categories. We report the mean and variance over 10 runs. Abbreviations: SC (Scene Caption), PE (Perception), TG (Temporal Grounding), UN (Uncertainty).}
\label{tab:small_category_reliability}
\vspace{-3mm}
\end{table}
\section{Conclusion}

WaymoQA reveals a clear gap between general vision-language ability and safety-critical driving reasoning. 
Pretrained MLLMs struggle when safety-critical scenes require connecting perception, prediction, and planning under action-induced downstream risks, highlighting the importance of reasoning about the consequences of ego actions.

Fine-tuning on WaymoQA consistently improves ImageQA, VideoQA, open-ended reasoning, and two-stage consistency, while multi-view ablations confirm the importance of surrounding context for safety-critical decisions. 
However, performance degradation under adverse conditions indicates that robust safety-critical reasoning remains an open challenge.

Overall, WaymoQA provides a 35K human-annotated multi-view benchmark and training resource that formulates safety-critical reasoning as immediate-risk resolution followed by downstream-risk mitigation.

\section*{Limitations}
WaymoQA has several limitations. 
\textbf{First}, WaymoQA is built from a single source, the Waymo End-to-End dataset, and therefore reflects its geographic, sensor, and scenario distributions.
Here, \emph{long-tail} refers to rare NHTSA pre-crash scenario types rather than rare environmental conditions.
As shown in our environmental analysis, common conditions such as bright daytime, dry roads, and light traffic are more frequent, while heavy traffic, low visibility, dark night, and adverse weather remain underrepresented.
This limits the diversity of safety-critical driving conditions covered by the benchmark.

\textbf{Second}, WaymoQA primarily evaluates high-level safety-critical reasoning rather than closed-loop driving control.
Although our planning-transfer and preliminary Bench2Drive results suggest that WaymoQA supervision can benefit downstream driving behavior, the closed-loop evaluation is limited in scale.
Larger-scale simulation and real-world validation are therefore needed before drawing conclusions about deployable driving safety.

\textbf{Third}, our main evaluation relies on human-written multiple-choice questions.
This design enables objective and reproducible scoring, but it may simplify the space of valid driving decisions and cannot fully capture open-ended safety-critical reasoning.
We therefore include open-ended evaluation as a complementary analysis, although automatic free-form metrics remain imperfect for assessing causal reasoning and action justification.

\textbf{Finally}, WaymoQA decomposes safety-critical reasoning into structured question categories.
While this enables diagnosis of failures in perception, prediction, planning, and uncertainty, it does not fully enforce a single end-to-end reasoning chain from visual evidence to final action.
Future work can extend WaymoQA toward compound multi-step reasoning with partial-credit evaluation.

\section*{Ethical Considerations}
\label{sup:ethics}

WaymoQA is built from the Waymo End-to-End dataset and follows the original Waymo data license and access terms. 
We will release annotations and code for research use, while raw images and videos must be accessed through the original Waymo dataset. 

The source Waymo E2E data are distributed under the privacy and access procedures of the original Waymo dataset. 
We do not redistribute raw images or videos, and users must obtain them through the original dataset release. 
Our released annotations consist of question-answer pairs, category labels, and evaluation metadata, and do not include personally identifying textual information. 
During annotation and verification, we also check that QA contents do not name or uniquely identify individual people and do not contain offensive or irrelevant content.

\section*{Acknowledgments}

This work was supported by the Basic Science Research Program through the National Research Foundation of Korea (NRF) funded by the Korea government (MSIT) (No. RS-2025-00520207); Institute of Information \& Communications Technology Planning \& Evaluation (IITP) grants funded by the Korea government (MSIT) (Nos. RS-2024-00457882, 2022-0-01045, 2022-0-00680, RS-2019-II190075); the Advanced GPU Utilization Support Program funded by the Government of the Republic of Korea (Ministry of Science and ICT); the AI Computing Infrastructure Enhancement (GPU Rental Support) User Support Program funded by the Ministry of Science and ICT (MSIT), Republic of Korea (No. RQT-25-120217); a grant partly supported by both IITP (MSIT) and Korea Evaluation Institute of Industrial Technology (KEIT) (MOTIE) (No. RS-2025-02217259); and the next-generation CultureTech technology development (R\&D) project for cultural space AX transformation through the Korea Creative Content Agency (KOCCA) grant funded by the Ministry of Culture, Sports and Tourism (MCST) in 2026 (No. RS-2026-25508598).

\bibliography{custom}

\clearpage
\appendix

\section*{Appendix Overview}
This appendix provides the following supplementary materials:

\begin{itemize}
    \item \textbf{Implementation details} (Appendix~\ref{sup:implementation}): LoRA setup, training hyperparameters, sampling strategy, and hardware configuration.
    \item \textbf{Dataset documentation} (Appendix~\ref{sup:details}): question-category statistics, task definitions, and representative templates.
    \item \textbf{MCQ construction} (Appendix~\ref{sup:mcq}): criteria for converting human annotations into four-option multiple-choice questions.
    \item \textbf{Complete benchmark tables} (Appendix~\ref{app:complete_tables}): full ImageQA and VideoQA results for all models and categories.
    \item \textbf{Additional ablations} (Appendix~\ref{sup:ablation}): training-modality ablations comparing Image-only, Video-only, and joint Image\&Video fine-tuning.
    \item \textbf{Open-ended evaluation} (Appendix~\ref{app:open_ended}): implementation details, prompts, package settings, and full GPTScore, BLEU, ROUGE, and BERTScore results across scene groups.
    \item \textbf{Environmental augmentation examples} (Appendix~\ref{app:augmentation_examples}): qualitative examples of appearance changes under dusk/dawn, foggy, nighttime, overcast, rainy, and snowy conditions.
    \item \textbf{Failure analysis} (Appendix~\ref{sup:fail}): a representative two-stage reasoning failure involving relative spatial directions.
    \item \textbf{Extending WaymoQA} (Appendix~\ref{sup:extending}): future extensions toward compound multi-step reasoning and partial-credit evaluation.
\end{itemize}

\section{Fine-Tuning and Implementation Details}
\label{sup:implementation}
\subsection{Fine-Tuning Details}
We fine-tune Qwen2.5 VL(7B) following the official guideline~\cite{Qwen2-VL-Finetuning}. 
The base weights of the language and vision towers and the projector are frozen, and LoRA adapters are applied to all supported modules except the LM head and token embeddings.
The batch size is 32 and learning rate is 2e-4. 
For data construction, videos are packed with all QA pairs associated with the clip and key frames are packed with all QA pairs anchored to the frame.

For training stability, we sample video at 1.2 FPS and train with a multi image fine tuning setup analogous to Image QA. 
Video data consist of 22 second multi view clips that impose substantial memory demand; accordingly, we use 1.2 FPS under the available hardware constraints. 
We expect non-uniform sampling to improve learning by allocating samples to key frames and fewer to less informative frames.
In practice the 28,585 training pool benefited from allocating at least 20,000 examples to training. 
Validation sets larger than 8,000 were associated with lower accuracy.

\subsection{Implementation Details}
\label{sup:implemeatation}
We fine-tune Qwen2.5-VL-7B-Instruct on WaymoQA using 8$\times$NVIDIA RTX A6000 GPUs with DeepSpeed ZeRO-2. 
We adopt parameter-efficient fine-tuning with LoRA following the official guideline~\cite{Qwen2-VL-Finetuning}. 
Both language and vision modules are equipped with LoRA adapters (rank 64, $\alpha=64$, dropout 0.05), while the base vision tower, language model, and merger module are frozen.
Following the official Qwen2.5-VL setup, we exclude the token embedding and output head from LoRA
(\texttt{lm\_head}, \texttt{embed\_tokens}) and keep all other linear layers trainable through adapters.
We enable Liger kernels for efficient attention, gradient checkpointing for memory savings, and train in \texttt{bfloat16} with \texttt{tf32} enabled on matrix multiplications.

Training uses a global batch size of 32 (per-device batch size 1 on 8 GPUs with gradient accumulation of 4), for 3 epochs over the WaymoQA training split.
We apply a cosine learning rate schedule with an initial learning rate of $2\times10^{-4}$,
a warm-up ratio of 0.03, and weight decay of 0.1.
Images are dynamically resized within the range of $256\times28\times28$ to $1024\times28\times28$ pixels to balance resolution and memory usage.

\section{Question Categories and Statistics}
\label{sup:details}

\begin{figure}[!h]
  \centering
  \includegraphics[width=0.9\linewidth]{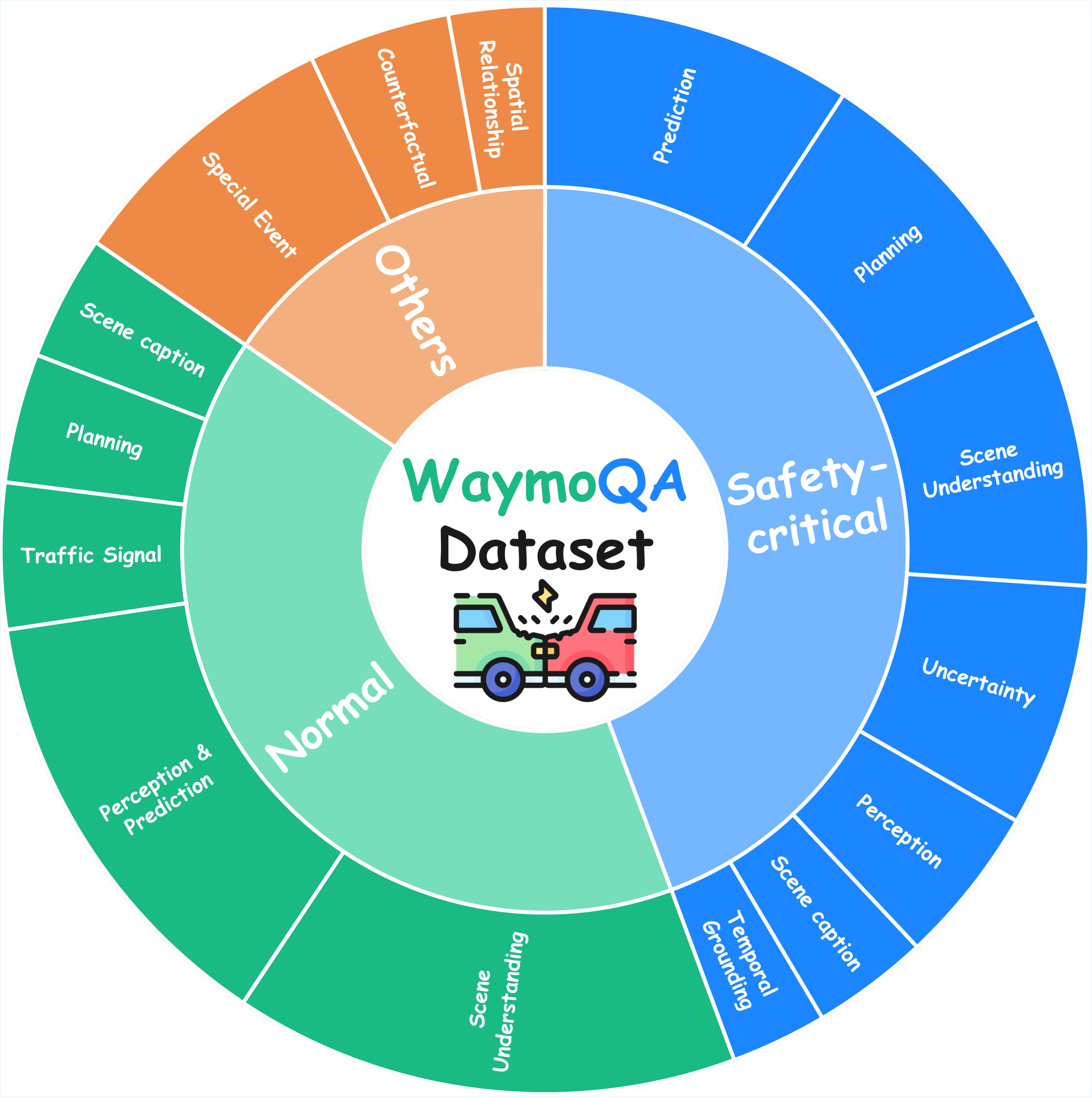}
  \caption{VQA distribution in the WaymoQA dataset.}
  \label{fig:sunburst}
\end{figure}

Table~\ref{tab:qb_stats} and Figure~\ref{fig:sunburst} summarizes how the 62-item question bank is instantiated across \textbf{Format} (Image / Video), \textbf{Scene type} (Normal / Safety-critical / Others), and \textbf{Task} categories.

\begin{table}[h]
\centering
\renewcommand{\arraystretch}{0.8}
\resizebox{\columnwidth}{!}{%
\begin{tabular}{l l l r}
\toprule
\textbf{Type} & \textbf{Scene} & \textbf{Category} & \textbf{Count} \\
\midrule
Image & Normal          & Normal Perception \& Prediction        & 4,004 \\
Image & Normal          & Normal Planning                         & 1,518  \\
Image & Normal          & Normal scene Understanding              & 4,606 \\
Image & Normal          & Normal scene caption                    & 1,134  \\
Image & Normal          & Traffic Signal                          & 1,266  \\
Image & Others          & Counterfactual                          & 1,616  \\
Image & Others          & Spatial Relationship                    & 351   \\
Image & Others          & Special Event                           & 1,927 \\
Image & Safety-critical & Safety-critical Perception              & 1,375  \\
Image & Safety-critical & Safety-critical Planning                & 3,239 \\
Image & Safety-critical & Safety-critical Prediction              & 2,990 \\
Image & Safety-critical & Safety-critical scene Understanding     & 1,926 \\
Image & Safety-critical & Safety-critical scene caption           & 1,060  \\
Image & Safety-critical & Uncertainty                             & 2,317 \\
\midrule
Video & Normal          & Normal Perception \& Prediction         & 1,080  \\
Video & Normal          & Normal scene Understanding              & 1,270  \\
Video & Normal          & Normal scene caption                    & 290  \\
Video & Normal          & Traffic Signal                          & 300  \\
Video & Others          & Counterfactual                          & 316  \\
Video & Others          & Spatial Relationship                    & 143   \\
Video & Others          & Special Event                           & 202  \\
Video & Safety-critical & Safety-critical Perception              & 324  \\
Video & Safety-critical & Safety-critical scene Understanding     & 874  \\
Video & Safety-critical & Safety-critical scene caption           & 290  \\
Video & Safety-critical & Temporal Grounding                      & 316  \\
Video & Safety-critical & Uncertainty                             & 288  \\
\midrule
\textbf{Sum} & & & 35,022\\
\bottomrule
\end{tabular}}
\caption{Distribution of questions in WaymoQA by format, scene type, and task category.}
\label{tab:qb_stats}
\end{table}

\noindent\textbf{Task definitions.}
Below we briefly describe each reasoning category covered by the question bank.\\

\begin{table*}[t]
\centering
\small
\renewcommand{\arraystretch}{0.95}
\resizebox{\textwidth}{!}{%
\begin{tabular}{p{0.22\textwidth} p{0.78\textwidth}}
\toprule
\textbf{Category} & \textbf{Representative question templates} \\
\midrule
Normal Prediction \& Planning
& Based on the current appearance of the person or two-wheeled vehicle, how do they intend to move? \newline
Which of the following describes a non-safety-critical object? \\
\midrule
Normal Planning
& What actions are currently available? \\
\midrule
Normal Scene Understanding
& Which of the following statements about traffic density is correct? \newline
How is the behavior of the vehicle summarized before the event? \\
\midrule
Normal Scene Caption
& Which of the following is correct as an image caption from the perspective of a normal frame? \newline
What is the video summary from the perspective of a normal scene? \\
\midrule
Counterfactual
& What situation might occur if the vehicle acted this way? \newline
What are the potential situations that could occur in the current scenario? \\
\midrule
Spatial Relationship
& What is the spatial relationship of two or more objects? \\
\midrule
Special Event
& Is there a special situation? \newline
What are the restrictions on behavior in school zones or congested areas? \newline
What is the plan for yielding when an ambulance or fire truck approaches? \\
\midrule
Safety-critical Perception
& What is a safety-critical object? \\
\midrule
Safety-critical Planning
& What are the driving guidelines for the ego vehicle based on the behavior of safety-critical objects? \newline
How should the ego vehicle drive to solve the first hazard while also addressing the second hazard? \\
\midrule
Safety-critical Prediction
& How do you think a safety-critical object will move? \newline
Are there any other safety-critical objects besides the first safety-critical object? If so, how would they move? \\
\midrule
Safety-critical Scene Understanding
& What is the correlation in safety-critical situations? \newline
How did the safety-critical object move and how did the ego vehicle behave? \\
\midrule
Safety-critical Scene Caption
& What is the correct image caption from the perspective of a safety-critical frame? \newline
What is correct regarding video summary from a safety-critical scene perspective? \\
\midrule
Uncertainty
& Where is the area that hides the danger in the current frame? \newline
Is there a blind spot in the field of view? \\
\midrule
Temporal Grounding
& Where is the safety-critical situation in the video? \\
\midrule
Traffic Signal
& Which of the following is a correct description of traffic signals? \\
\bottomrule
\end{tabular}}
\caption{Representative question templates for each reasoning category in the WaymoQA question bank.}
\label{tab:qb_templates}
\end{table*}

\noindent\textbf{Caption.}
These questions ask the model to produce or select a caption summarizing the scene.
For \emph{normal} scenes, the caption focuses on overall driving context such as road layout, traffic participants, and general situation.
For \emph{safety critical} scenes, the caption is required to highlight the safety relevant situation, such as an imminent collision risk or a blocked lane.\\

\noindent\textbf{Scene Understanding.}
Compared to Caption, these questions probe more fine grained comprehension of the scene.
They query specific aspects of the environment such as which lanes are drivable, who has right of way, and where the ego vehicle is positioned rather than a single high level summary.\\

\noindent\textbf{Perception.}
Perception questions ask about properties or states of objects in the scene:
their type such as car, pedestrian, cyclist, or construction vehicle, their attributes such as parked or moving or indicating, and their positions relative to the ego vehicle or road structure.\\

\noindent\textbf{Prediction.}
Prediction questions require anticipating the \emph{future behavior} of agents.
Typical queries include whether another vehicle will cut in, whether a pedestrian is likely to cross, or how surrounding traffic will evolve in the next few seconds.\\

\noindent\textbf{Planning.}
Planning questions focus on the ego vehicle’s \emph{future action}.
The model must decide an appropriate maneuver such as continue, slow down, stop, or change lanes, given the perceived scene and expected motions of other agents.\\

\noindent\textbf{Uncertainty.}
These questions target reasoning about unknown or unobserved risk, such as occluded areas, blind spots, or partially visible agents.
The model must recognize where it cannot be certain, for example when a pedestrian may emerge from behind a truck, and reflect this uncertainty in its choice.\\

\noindent\textbf{Counterfactual.}
Counterfactual questions describe hypothetical deviations from the current situation.
They typically ask what would happen \emph{if} the ego vehicle or another agent took a different action, for example what would likely occur if the ego vehicle changed lanes now, probing causal reasoning about alternative behaviors.\\

\noindent\textbf{Special Event.}
Special Event questions focus on unusual but practically important conditions, such as road construction, emergency vehicles, accidents, or adverse weather such as rain, snow, or poor visibility.
The model must detect these events and understand their impact on safe driving.\\

\noindent\textbf{Spatial Relationship.}
These questions query the interaction and relative positioning between two or more objects.
Examples include determining which vehicle is ahead or behind, which lane another car occupies, or whether two agents are on a collision course.\\

\noindent\textbf{Temporal Grounding.}
Temporal Grounding questions ask \emph{when} in a video the situation becomes safety critical.
The model must identify the time interval or frame range where risk emerges or peaks, rather than only describing a single static frame, and this category is defined for video data only.\\

\noindent\textbf{Traffic Signal.}
These questions assess correct interpretation of traffic lights, arrows, and lane markings and how they constrain ego behavior.
They often require relating signal state such as red light, protected left arrow, or lane only sign to valid or invalid driving actions.\\

We further provide representative question templates for each reasoning category in Table~\ref{tab:qb_templates}.

\section{Criteria of Converting Annotation into MCQ}
\label{sup:mcq}

In order to construct the multiple-choice questions (MCQ) used in WaymoQA, we convert human-annotated answers into a structured four-choice format. Each MCQ requires one correct answer and three plausible false candidates. The false answers must be carefully designed so that the overall difficulty remains appropriate. Our target difficulty range ensures that pretrained models score roughly between 30\% and 70\% because scores below 30\% indicate that the question is unnecessarily difficult, and scores above 70\% mean that the question is too easy. We iteratively refine the criteria by creating subsets of MCQs, and adjusting the false-choices construction rules. The final criteria are summarized below.\\

\noindent
\textbf{1. The question must not be solvable from text input alone.}
False choices that are irrelevant, exaggerated, or logically disconnected allow models to answer correctly without looking at the image. To avoid this issue, all candidates must remain close to the topic of the question. Each false answer should carry plausible content that resembles the correct answer so that visual understanding is required.\\

\noindent
\textbf{2. False choices must be visually similar to the correct answer.}
To ensure that the MCQ reflects genuine visual reasoning, false candidates must be close variants of the correct answer. We modify only minor attributes such as object type, trajectory, color, count, or location. These changes create subtle perturbations that require precise image understanding.
For example:
\begin{itemize}
\item \textbf{True}: \textit{``A \underline{black} vehicle is \underline{going straight}."} \\→ \textbf{False}: \textit{``A \underline{white} vehicle is \underline{turning right}."}  
\item \textbf{True}: \textit{``\underline{Two} cars are parked on the \underline{right} side of the road."} \\→ \textbf{False}: \textit{``\underline{Five} cars are parked on the \underline{left} side of the road."}\\
\end{itemize}

\noindent
\textbf{3. Each question must have exactly one correct answer.}
We manually verify that no false candidate can be interpreted as correct. If a candidate introduces ambiguity or can be considered a valid answer under certain contexts, we revise or discard it.\\

\noindent
\textbf{4. All choices must follow a consistent writing style.}
Since multiple annotators contribute to the dataset, differences in tone and phrasing naturally occur. A final reviewer rewrites all answer options to unify the style across the dataset. This process ensures that sentence structure, wording, and perspective remain consistent across all choices.\\

\section{Complete Benchmark Tables}
\label{app:complete_tables}

The main paper reports a compact table that merges ImageQA and VideoQA. Here we provide the original detailed ImageQA and VideoQA tables before merging. Table~\ref{tab:imageqa} and ~\ref{tab:videoqa} show all Normal, Others, Safety-Critical, and Overall category scores.
\begin{table*}[t]
\centering
\resizebox{\textwidth}{!}{%
\begin{tabular}{l | *{6}{c} | *{3}{c} | *{7}{S} | c}
\toprule
\multicolumn{1}{l|}{\textbf{Model}} &
\multicolumn{6}{c|}{\textbf{Normal Scene}} &
\multicolumn{3}{c|}{\textbf{Others}} &
\multicolumn{7}{c|}{\cellcolor{SChead}\textbf{Safety - Critical}} &
\multicolumn{1}{c}{\textbf{Overall}}\\
\cmidrule(lr){2-7}\cmidrule(lr){8-10}\cmidrule(lr){11-17}
 & TS & SC & SU & PL & PP & Overall
 & CF & SE & Overall
 & SU & PE & PR & PL & SC & UN & Overall
 &  \\
\midrule
Human Experts & 96.6 & 97.7 & 97.7 & 95.3 & 94.0 & 96.1 & 94.8 & 96.7 & 96.3 & 96.8 & 98.0 & 96.4 & 94.3 & 98.7 & 94.9 & 96.0 & 96.1\\
GPT 5.2 & 82.0 & 88.9 & 73.1 & 71.7 & 75.4 & 75.7 & 56.4 & 57.1 & 57.0 & 68.4 & 65.6 & 73.8 & 62.3 & 57.6 & 53.0 & 64.5 & 67.4 \\
\midrule
LLaVA1.5(7B)              & 19.6 & 24.4 & 50.0 & 24.4 & 32.8 & 35.9 & 30.7 & 39.0 & 37.5 & 24.5 & 21.3 & 24.7 & 25.5 & 27.2 & 17.0 & 23.2 & 30.7 \\
LLaVA-OneVision(7B)       & 71.3 & \underline{78.5} & \textbf{73.3} & \underline{65.5} & 56.1 & 66.7 & \underline{53.6} & 51.4 & 51.7 & 50.8 & 42.2 & 56.9 & 53.2 & 45.5 & \textbf{64.9} & 54.4 & 58.6 \\
InternVL3.5(8B)           & 14.7 & 31.8 & 45.3 & 60.8 & 37.4 & 40.1 & 41.3 & 47.9 & 46.6 & 26.6 & 30.4 & 23.7 & 41.1 & 25.3 & 22.2 & 28.9 & 37.0 \\
InternVL3.5(4B)           & 17.6 & 32.5 & 42.5 & 43.0 & 39.6 & 38.0 & 40.2 & 42.8 & 42.3 & 21.7 & 24.1 & 23.6 & 39.2 & 41.1 & 20.2 & 27.8 & 34.4 \\
InternVL3(8B)             & 16.8 & 29.6 & 43.4 & 46.8 & 24.8 & 33.6 & 34.0 & 49.9 & 47.0 & 28.8 & 24.9 & 27.1 & 46.2 & 41.7 & 27.8 & 32.9 & 35.9 \\
InternVL3(14B)            & 20.2 & 31.2 & 45.9 & 50.0 & 39.2 & 40.2 & 42.1 & 58.6 & 55.6 & 23.4 & 32.8 & 52.1 & 45.3 & 45.8 & 39.0 & 42.1 & 44.0 \\
Qwen2.5-VL(32B) & 79.9 & \textbf{81.1} & 70.7 & 64.4 & \underline{69.4} & \underline{71.2} & 50.0 & 62.6 & 60.3 & 52.6 & 44.6 & \underline{74.3} & \underline{64.6} & \underline{56.4} & 61.4 & \underline{62.6} & \underline{65.4}\\
Qwen2.5-VL(7B)            & 77.4 & 77.4 & 71.2 & 48.4 & 63.3 & 68.1 & 43.5 & 58.9 & 55.6 & 54.4 & 55.3 & 71.0 & 64.0 & 43.0 & 54.6 & 61.1 & 62.4 \\
Phi3.5-vision & 24.1 & 32.5 & 63.2 & 61.2 & 40.6 & 48.9 & 44.6 & 44.5 & 44.5 & 32.0 & 25.6 & 41.8 & 45.6 & 44.3 & 46.5 & 40.9 & 44.7 \\
Gemma3(4B) & 60.2 & 62.6 & 62.6 & 56.2 & 57.1 & 59.7 & 48.0 & 48.9 & 48.7 & 34.2 & 37.1 & 53.8 & 44.3 & 51.3 & 33.8 & 43.5 & 50.6\\
Gemma3(27B) & 67.6 & 73.3 & 68.7 & 51.5 & 61.5 & 64.3 & 49.1 & \underline{65.6} & \underline{62.6} & \underline{67.2} & \underline{61.6} & 67.1 & 59.4 & 54.4 & 49.4 & 60.4 & 62.3\\
\midrule
InternVL2-DA-DriveLM(4B)  & 34.0 & 58.5 & 60.2 & \textbf{66.2} & 49.6 & 54.2 & 34.6 & 38.5 & 37.7 & 44.4 & 24.1 & 41.2 & 26.8 & 38.6 & 46.1 & 36.8 & 43.6\\
OpendriveVLA & \underline{82.0} & 76.0 & 42.1 & 50.3 & 57.1 & 54.9  & 41.3 & 44.5 & 43.9 & 57.1 & 50.7 & 68.3 & 53.1 & 41.1 & 49.6 & 56.0 & 53.4\\
\midrule
Qwen2.5-VL(7B) (FT, Ours)       & \textbf{84.4} & 72.6 & \underline{72.0} & 58.5 & \textbf{80.6} & \textbf{74.9} & \textbf{63.1} & \textbf{74.8} & \textbf{72.7} & \textbf{76.1} & \textbf{76.6} & \textbf{77.8} & \textbf{65.9} & \textbf{67.8} & \underline{63.8} & \textbf{71.2} & \textbf{72.9} \\
\bottomrule
\end{tabular}}
\caption{Detailed Image QA performance of the MLLMs on WaymoQA. \textbf{Bold}: best; \underline{underline}: second best. The following abbreviations are used:
TS (Traffic Signal), SC (Scene Caption), SU (Scene Understanding), PL (Planning), PE (Perception), PR (Prediction), PP (Perception\&Prediction), CF (Counterfactual), SE (Special Event), UN (Uncertainty), FT (Fine-tuning).}
\label{tab:imageqa}
\end{table*}
\begin{table*}[t]
\centering
\resizebox{\textwidth}{!}{%
\begin{tabular}{l | *{5}{c} | *{4}{c} | *{6}{S} | c}
\toprule
\multicolumn{1}{l|}{\textbf{Model}}
& \multicolumn{5}{c|}{\textbf{Normal Scene}}
& \multicolumn{4}{c|}{\textbf{Others}}
& \multicolumn{6}{>{\columncolor{SChead}}c|}{\textbf{Safety - Critical}}
& \multicolumn{1}{c}{\textbf{Overall}} \\
\cmidrule(lr){2-6}\cmidrule(lr){7-10}\cmidrule(lr){11-16}
& TS & SC & SU & PP & Overall
& CF & SE & SR & Overall
& SC & SU & PE & TG & UN & Overall
& \\
\midrule
Human Experts                    & 95.4 & 95.7 & 100.0 & 98.7 & 96.9 & 95.9 & 100.0 & 90.5 & 96.6 & 97.2 & 97.2 & 96.0 & 94.0 & 92.5 & 95.9 & 96.5 \\
GPT 5.2 & 74.0 & 70.6 & 76.3 & 85.3 & 79.1 & 82.0 & 89.4 & 47.6 & 78.8 & 54.1 & 65.5 & 50.0 & 49.0 & 81.3 & 61.6 & 72.6 \\
\midrule
LLaVA1.5(7B)                 & 26.0 & 29.4 & 48.8 & 40.4 & 42.4 & 30.0 & 40.4 & 33.3 & 34.7 & 13.5 & 27.5 & 32.6 & 27.4 & 33.3 & 27.5 & 35.9 \\
LLaVA-OneVision(7B)          & \underline{80.0} & 35.2 & 68.8 & 60.7 & 65.8 & \textbf{80.0} & \underline{93.6} & \textbf{61.9} & \underline{82.1} & 32.4 & 43.4 & 40.3 & 29.4 & 62.5 & 42.3 & 59.2 \\
InternVL3.5(8B)              & 30.0 & 17.6 & 48.8 & 47.8 & 45.1 & 52.0 & 61.7 & 23.8 & 50.8 & 29.7 & 28.2 & 51.9 & 19.6 & 43.7 & 32.9 & 41.4 \\
InternVL3.5(4B)              & 26.0 & 41.1 & 44.1 & 28.8 & 36.3 & 62.0 & 63.8 & 38.0 & 58.4 & 37.8 & 23.4 & 42.3 & 17.6 & 39.5 & 29.3 & 36.7 \\
InternVL3(8B)                & 26.0 & 35.2 & 38.1 & 37.4 & 36.3 & 58.0 & 29.7 & 19.0 & 39.7 & 40.5 & 27.5 & 50.0 & 25.4 & 56.2 & 36.2 & 36.8 \\
InternVL3(14B)               & 28.5 & 50.0 & 51.2 & 46.0 & 46.6 & \underline{72.0} & 60.4 & 33.3 & 60.4 & \underline{59.4} & 27.3 & 51.9 & 25.0 & 57.4 & 38.6 & 46.3 \\
Qwen2.5-VL(32B) & \underline{80.0} & 52.9 & \underline{70.6} & 74.2 & 72.3 & 76.6 &93.1 & \underline{57.3} &79.7  & 43.2 & 46.2 & 51.9 & 31.3 & \underline{67.8} & 47.6  & \underline{64.1}\\
Qwen2.5-VL(7B)               & \underline{80.0} & 41.1 & \underline{70.6} & \underline{76.6} & \underline{72.7} & 48.0 & 89.2 & 52.3 & 64.4 & 32.4 & \underline{53.1} & \underline{55.7} & 31.3 & 64.5 & 49.5 & 63.8 \\
Phi3.5-vision & 35.0 & 29.4 & 62.3 & 43.2 & 50.9 & 67.0 & 40.4 & 45.2 & 52.5 & 37.8 & 31.7 & 32.6 & 24.5 & 54.1 & 32.6 & 45.1 \\
Gemma3(4B) & 58.0 & 52.9 & 62.3 & 58.3 & 60.0 & 70.0 & 76.6 & 33.3 & 66.1 & 54.0 & 29.6 & 38.5 & \underline{54.9} & 54.2 & 41.1 & 53.8\\
Gemma3(27B) & 74.0 & 47.1 & 67.4 & 61.3 & 65.1 & 66.0 & 82.9 & 47.6 & 69.4 & 43.2 & 53.7 & 55.7 & 39.2 & 62.5 & \underline{51.8} & 60.7\\
\midrule
InternVL2-DA-DriveLM(4B) & 50.0 & \textbf{70.5} & 57.2 & 57.6 & 57.0 & 36.0 & 70.2 & 52.3 & 52.5 & 37.8 & 43.4 & 30.7 & 27.4 & 58.3 & 40.5 & 50.3 \\
OpendriveVLA  & 74.0 & 52.9 & 62.3 & 75.1 & 67.9 & \underline{72.0} & 89.4 & 33.3 & 72.0 & 43.2 & 49.7 & 50.0 & 35.3 & 54.1 & 47.5 & 60.9 \\
\midrule
Qwen2.5-VL(7B) (FT, Ours)  & \textbf{84.0} & \underline{64.7} & \textbf{70.7} & \textbf{81.6} & \textbf{76.0} & \textbf{80.0} & \textbf{95.8} & \textbf{61.9} & \textbf{83.1} & \textbf{63.5} & \textbf{66.2} & \textbf{63.5} & \textbf{58.9} & \textbf{70.8} & \textbf{65.0} & \textbf{72.8} \\
\bottomrule
\end{tabular}%
}
\caption{Detailed Video QA performance of the MLLMs on WaymoQA. \textbf{Bold}: best; \underline{underline}: second best. The following abbreviations are used:
TS (Traffic Signal), SC (Scene Caption), SU (Scene Understanding), PL (Planning), PE (Perception), PP (Perception\&Prediction), CF (Counterfactual), SE (Special Event), SR (Spatial Relationship), TG (Temporal Grounding), UN (Uncertainty), FT (Fine-tuning).}
\label{tab:videoqa}
\end{table*}

\section{Ablation Study}
\label{sup:ablation}
\begin{table*}[t]
\centering
\small                                      
\setlength{\tabcolsep}{2.5pt}               
\renewcommand{\arraystretch}{1.05}          
\resizebox{\textwidth}{!}{%
\begin{tabular}{l | c | *{6}{c} | *{3}{c} | *{7}{S} | c}
\toprule
\multicolumn{19}{c}{\textbf{Image QA}}\\
\midrule
\multicolumn{1}{l|}{\textbf{Model}} &
\multicolumn{1}{c|}{\textbf{Train Dataset}} &
\multicolumn{6}{c|}{\textbf{Normal Scene}} &
\multicolumn{3}{c|}{\textbf{Others}} &
\multicolumn{7}{c|}{\cellcolor{SChead}\textbf{Safety - Critical}} &
\multicolumn{1}{c}{\textbf{Overall}}\\
\cmidrule(lr){3-8}\cmidrule(lr){9-11}\cmidrule(lr){12-18}
& & TS & SC & SU & PL & PP & Overall
 & CF & SE & Overall
 & SU & PE & PR & PL & SC & UN & Overall
 &  \\
\midrule
Random Guess &- & 25.0 & 25.0& 25.0& 25.0& 25.0& 25.0& 25.0& 25.0& 25.0& 25.0& 25.0& 25.0& 25.0& 25.0& 25.0 & 25.0& 25.0\\
Human Experts &- & 96.6 & 97.7 & 97.7 & 95.3 & 94.0 & 96.1 & 94.8 & 96.7 & 96.3 & 96.8 & 98.0 & 96.4 & 94.3 & 98.7 & 94.9 & 96.0 & 96.1\\
\midrule
Qwen2.5-VL(7B)~\cite{bai2025qwen2} (Baseline)        &-   & 77.4 & 77.4 & 71.2 & 48.4 & 63.3 & 68.1 & 43.5 & 58.9 & 55.6 & 54.4 & 55.3 & 71.0 & 64.0 & 43.0 & 54.6 & 61.1 & 62.4 \\
\midrule
Qwen2.5-VL(7B)~\cite{bai2025qwen2} (FT, Ours) &  Image\&Video  & \textbf{84.4} & 72.6 & \underline{72.0} & \textbf{58.5} & \textbf{80.6} & \textbf{74.9} & \textbf{63.1} & \textbf{74.8} & \textbf{72.7} & \textbf{76.1} & \textbf{76.6} & \textbf{77.8} & \textbf{65.9} & \textbf{67.8} & \textbf{63.8} & \textbf{71.2} & \textbf{72.9} \\
Qwen2.5-VL(7B)~\cite{bai2025qwen2} (FT, Ours) &  Image  & \underline{79.1} & \underline{78.5} & \textbf{73.6} & \underline{53.9} & \underline{71.1} & \underline{71.3} & \underline{49.1} & \underline{61.5} & \underline{59.3} & \underline{61.6} & \underline{64.0} & \underline{73.9} & \underline{64.9} & \underline{44.3} & 54.3 & \underline{63.6} & \underline{65.7} \\
Qwen2.5-VL(7B)~\cite{bai2025qwen2} (FT, Ours) &  Video  & 78.6 & \textbf{80.0} & 71.6 & 50.4 & 66.7 & 68.7 & 46.4 & 59.1 & 56.8 & 59.4 & 59.3 & 70.9 & 62.0 & 43.7 & \underline{54.8} & 61.3 & 63.4\\
\bottomrule
\end{tabular}}

\setlength{\tabcolsep}{2.5pt}              
\renewcommand{\arraystretch}{1.05}
\resizebox{\textwidth}{!}{%
\begin{tabular}{l | c | *{5}{c} | *{4}{c} | *{6}{S} | c}
\toprule
\multicolumn{18}{c}{\textbf{Video QA}}\\
\midrule
\multicolumn{1}{l|}{\textbf{Model}} &
\multicolumn{1}{c|}{\textbf{Train Dataset}} &
\multicolumn{5}{c|}{\textbf{Normal Scene}} &
\multicolumn{4}{c|}{\textbf{Others}} &
\multicolumn{6}{>{\columncolor{SChead}}c|}{\textbf{Safety - Critical}} &
\multicolumn{1}{c}{\textbf{Overall}} \\
\cmidrule(lr){3-7}\cmidrule(lr){8-11}\cmidrule(lr){12-17}
& & TS & SC & SU & PP & Overall
& CF & SE & SR & Overall
& SC & SU & PE & TG & UN & Overall
& \\
\midrule
Random Guess &- & 25.0& 25.0& 25.0& 25.0& 25.0& 25.0& 25.0& 25.0& 25.0& 25.0& 25.0& 25.0& 25.0& 25.0& 25.0& 25.0\\
Human Experts            &   -     & 95.4 & 95.7 & 100.0 & 98.7 & 96.9 & 95.9 & 100.0 & 90.5 & 96.6 & 97.2 & 97.2 & 96.0 & 94.0 & 92.5 & 95.9 & 96.5 \\
\midrule
Qwen2.5-VL(7B)~\cite{bai2025qwen2} (Baseline) &  -  & 80.0 & \underline{41.1} & \underline{70.6} & 76.6 & \underline{72.7} & 48.0 & 89.2 & 52.3 & 64.4 & 32.4 & 53.1 & 55.7 & 31.3 & 64.5 & 49.5 & 63.8 \\
\midrule
Qwen2.5-VL(7B)~\cite{bai2025qwen2} (FT, Ours)  & Image\&Video& \textbf{84.0} & \textbf{64.7} & \textbf{70.7} & \textbf{81.6} & \textbf{76.0} & \textbf{80.0} & \textbf{95.8} & \textbf{61.9} & \textbf{83.1} & \textbf{63.5} & \textbf{66.2} & \textbf{63.5} & \textbf{58.9} & \underline{70.8} & \textbf{65.0} & \textbf{72.8} \\
Qwen2.5-VL(7B)~\cite{bai2025qwen2} (FT, Ours) &  Image  & 78.0 & 35.3 & 68.8 & 76.7 & 71.4 & 48.0 & \underline{91.5} & 52.4 & 66.1 & 33.8 & \underline{58.0} & 51.0 & 36.3 & \textbf{71.9} & 52.9 & 63.9\\
Qwen2.5-VL(7B)~\cite{bai2025qwen2} (FT, Ours) &  Video  & \underline{81.0} & \underline{41.1} & \underline{69.1} & \underline{77.9} & 72.6 & \underline{51.0} & \underline{91.5} & \underline{54.8} & \underline{67.8} & \underline{37.9} & 56.6 & \underline{59.6} & \underline{39.2} & 68.8 & \underline{54.1} & \underline{65.1}\\
\bottomrule
\end{tabular}}
\caption{Ablation Study on Image and Video QA of WaymoQA. \textbf{Bold}: best; \underline{underline}: second best. The following abbreviations are used:
TS (Traffic Signal), SC (Scene Caption), SU (Scene Understanding), PL (Planning), PE (Perception), PR (Prediction), PP (Perception\&Prediction), CF (Counterfactual), SE (Special Event), SR (Spatial Relationship), TG (Temporal Grounding), UN (Uncertainty), FT (Fine-tuning).}
\label{tab:ablation}

\end{table*}
\paragraph{Ablation on training modality.}
Table~\ref{tab:ablation} analyzes how the choice of training modality affects performance on WaymoQA.  
Starting from the off-the-shelf Qwen2.5-VL(7B) baseline, we fine-tune three variants:
(i) on both Image and Video QA data,  
(ii) on Image QA only, and  
(iii) on Video QA only.

For \textbf{Image QA}, all fine-tuned models outperform the baseline (62.4\% overall), but the joint Image\&Video model is clearly the most effective.  
Training only on images raises overall Image QA accuracy to 65.7\%, while training only on videos yields 63.4\%.  
In contrast, using both modalities pushes Image QA to 72.9\% overall and improves every scene type: normal scenes from 68.1\% to 74.9\%, ``Others'' from 55.6\% to 72.7\%, and safety-critical scenes from 61.1\% to 71.2\%.  
Thus, even for image-only evaluation, exposure to video questions provides a substantial gain, especially on safety-critical categories.

A similar trend appears for \textbf{Video QA}.  
Fine-tuning on video-only data improves the baseline from 62.6\% to 65.1\% overall, and image-only training gives a smaller boost to 63.9\%.  
However, the Image\&Video model again achieves the best result with 72.8\% overall accuracy, lifting normal scenes from 72.7\% to 76.0\%, ``Others'' from 65.2\% to 83.1\%, and safety-critical scenes from 51.3\% to 65.0\%.  

These results confirm our hypothesis from the main text: while modality-specific fine-tuning helps the corresponding evaluation modality, jointly training on both image and video questions yields the largest improvements, particularly for safety-critical reasoning, by encouraging the model to share representations across complementary temporal and spatial cues.

\section{Open-ended Evaluation}
\label{app:open_ended}

In the main paper, we report a compact open-ended evaluation to complement the primary MCQ benchmark. 
Here, we provide the full open-ended results for each scene group: Normal Scene, Others, Safety-Critical Scene, and Overall. 
This analysis verifies whether the gains observed in the MCQ protocol also transfer to free-form answer generation.

We evaluate representative models using GPT-based scoring, BLEU-1/2/3/4, ROUGE-1/2/L, and BERTScore. 
Across all scene groups, the WaymoQA fine-tuned Qwen2.5-VL(7B) model achieves the strongest performance on GPTScore and most lexical/semantic metrics. 
This confirms that WaymoQA supervision improves not only answer-option selection, but also open-ended safety-critical reasoning. 
At the same time, the relative ordering among pretrained models is less consistent than in the MCQ benchmark, supporting our use of human-written MCQs as the primary evaluation protocol and open-ended scoring as complementary evidence.

\subsection{Evaluation Pipeline}

We implement open-ended evaluation as a three-stage pipeline. 
First, the evaluated VLM generates a free-form answer from the visual input and question. 
Second, a GPT-based judge assigns a semantic correctness score to the generated answer with respect to the human-written reference answer. 
Third, we compute automatic text metrics in batch and aggregate the results by scene group.

For open-ended generation, the model receives only the visual input and the question. 
The multiple-choice answer options are not provided to the model. 
This prevents open-ended evaluation from becoming answer-option selection and instead tests whether the model can produce the answer in natural language. 
For examples originally stored in MCQ format, we recover the textual reference by selecting the ground-truth option and removing the option prefix such as ``A.'' or ``B.''.

Input images are resized so that the maximum side length is at most 1024 pixels and encoded as JPEG with quality 85 before being passed to the model. 
The camera order follows a fixed priority order: FRONT, FRONT\_LEFT, FRONT\_RIGHT, REAR\_LEFT, REAR\_RIGHT, REAR, SIDE\_LEFT, and SIDE\_RIGHT. 
For generation, we use temperature $0.0$, top-$p=1.0$, and a maximum generation length of 64 tokens.

\subsection{GPT-based Scoring}

We use a GPT-based judge to assess semantic correctness beyond exact lexical overlap. 
The judge is instructed to act as an expert evaluator for autonomous-driving perception and planning. 
Each prediction is scored on a 0--100 scale using three criteria: factual correctness, completeness, and clarity/relevance. 
Factual correctness receives 50 points, completeness receives 30 points, and clarity/relevance receives 20 points. 
This weighting emphasizes whether the generated answer captures the correct driving fact, object state, or ego maneuver, which is especially important for safety-critical perception and planning questions.

The judge uses deterministic decoding with temperature $0.0$ and a maximum output length of 4 tokens. 
It is required to return only a single integer between 0 and 100. 
For planning questions, correctness of the final maneuver is treated as the most important factor. 
Hallucinated objects or events are penalized, and verbose answers are not rewarded when the core driving fact is incorrect.

\subsection{Automatic Text Metrics}

We compute BLEU-1/2/3/4, ROUGE-1/2/L, and BERTScore for open-ended evaluation. 
BLEU is computed with NLTK sentence-level BLEU using smoothing method 1. 
ROUGE is computed with the \texttt{rouge-score} package, where we report ROUGE-1, ROUGE-2, and ROUGE-L F1 scores with stemming enabled. 
BERTScore is computed with \texttt{bert-score} using \texttt{roberta-large} with \texttt{lang=en} and batch size 32. 
All metrics are computed at the sample level and averaged within each scene group. 
We use default package settings unless otherwise specified.

\subsection{Prompts}

The following system prompt is used for open-ended VLM generation.

\begin{lstlisting}[style=promptstyle, caption={System prompt for open-ended VLM generation.}, label={lst:vlm_open_prompt}]
You are an autonomous-driving scene question-answering assistant.
You will be given one or more images from vehicle cameras.
Answer the question concisely and accurately in 1-2 short sentences.
Do NOT list or reference multiple-choice options.
Base your answer only on what is clearly visible in the image(s).
If you are uncertain, state the most likely answer with a brief reason.
\end{lstlisting}

The following system prompt is used for GPT-based scoring.

\begin{lstlisting}[style=promptstyle, caption={System prompt for GPT-based open-ended answer scoring.}, label={lst:gpt_judge_prompt}]
You are an expert evaluator specializing in autonomous-driving perception and planning.
Your task is to score a model-generated answer against the ground-truth answer
for a driving scene question. Score on a 0-100 scale using the rubric below.

EVALUATION RUBRIC (total 100 points)

[A] Factual Correctness (0-50 pts)
50 : The predicted action / fact is exactly correct and matches the ground truth.
40 : Correct core action, minor wording or detail difference (e.g. "slow down" vs "decelerate").
25 : Partially correct -- mentions one correct element among multiple, or correct direction but wrong maneuver type.
10 : Related to the driving context but the specific action/fact is wrong.
 0 : Completely wrong (e.g., "turn right" when GT is "go straight").

[B] Completeness (0-30 pts)
30 : All key information from the ground truth is present in the prediction.
20 : Most key information present; one minor element is missing or imprecise.
10 : Only about half of the key information is captured.
 0 : Critical information (e.g., the main action) is missing.

[C] Clarity & Relevance (0-20 pts)
20 : Concise, on-topic, directly answers the question without irrelevant content.
12 : Mostly relevant but includes unnecessary filler or verbose phrasing.
 6 : Partially on-topic; the answer is confusing or mixes correct/incorrect elements.
 0 : Off-topic, incoherent, or refuses to answer.

SCORING NOTES
- Penalise hallucination (adding objects/events not in GT) in section [B].
- Reward semantic equivalence even when exact wording differs.
- For planning questions, correctness of the final decision matters most.
- Do NOT reward verbose answers that pad around a wrong core answer.

Respond with ONLY a single integer between 0 and 100. No explanation, no prefix.
\end{lstlisting}

For each example, the user prompt to the GPT judge contains the question, the ground-truth answer, and the predicted answer.

\begin{lstlisting}[style=promptstyle, caption={User prompt template for GPT-based open-ended answer scoring.}, label={lst:gpt_judge_user_prompt}]
[Question]
{question}

[Ground Truth Answer]
{gt_text}

[Predicted Answer]
{pred_text}

Score (0-100):
\end{lstlisting}

\subsection{Results}

As shown in Table~\ref{tab:supp_open_ended_full}, the WaymoQA fine-tuned Qwen2.5-VL(7B) model achieves the strongest performance across all scene groups on GPTScore and most lexical/semantic metrics. 
For example, on Overall questions, the fine-tuned model improves GPTScore from 25.5 to 40.4 over the Qwen2.5-VL(7B) base model, and also improves BLEU-4, ROUGE-L, and BERTScore. 
On Safety-Critical questions, the same model achieves the highest GPTScore and ROUGE-L, indicating that WaymoQA supervision improves free-form safety-critical reasoning rather than only answer-option selection.

At the same time, the relative ordering among pretrained models is less consistent than in the MCQ benchmark. 
Some models obtain relatively high lexical-overlap scores despite lower GPTScore, while others show similar BERTScore values despite different levels of driving correctness. 
This supports our evaluation design: human-written MCQs provide the primary objective benchmark, while open-ended evaluation serves as complementary evidence that the observed gains transfer to free-form generation.

\begin{table*}[t]
\centering
\setlength{\tabcolsep}{3.5pt}
\renewcommand{\arraystretch}{1.05}
\resizebox{0.92\textwidth}{!}{%
\begin{tabular}{llccccccccc}
\toprule
Scene Group & Model & GPTScore & B-1 & B-2 & B-3 & B-4 & R-1 & R-2 & R-L & BERTScore \\
\midrule
\multirow{7}{*}{Normal}
& LLaVA-OneVision(7B) & 26.5 & 8.89 & 4.74 & 3.27 & 2.41 & 14.3 & 4.17 & 12.8 & 85.3 \\
& InternVL3.5(8B) & 30.5 & 11.2 & 5.41 & 3.22 & 2.05 & 14.5 & 4.48 & 11.9 & 84.6 \\
& Qwen2.5-VL(7B) & 30.5 & 11.2 & 5.25 & 3.23 & 2.29 & 14.5 & 4.48 & 11.9 & 84.7 \\
& Qwen2.5-VL(32B) & 32.4 & 10.5 & 5.12 & 3.22 & 2.38 & 14.0 & 4.52 & 11.4 & 84.1 \\
& Gemma3(27B) & 32.1 & 11.8 & 5.24 & 2.89 & 1.86 & 15.2 & 3.85 & 12.4 & 84.7 \\
& InternVL2-DA-DriveLM(4B) & 28.0 & 12.2 & 6.82 & 4.95 & 3.81 & 18.4 & 6.04 & 16.5 & 86.5 \\
& Qwen2.5-VL(7B) (FT) & \textbf{45.7} & \textbf{28.9} & \textbf{18.6} & \textbf{13.7} & \textbf{11.4} & \textbf{39.3} & \textbf{18.1} & \textbf{36.7} & \textbf{89.2} \\
\midrule
\multirow{7}{*}{Others}
& LLaVA-OneVision(7B) & 21.7 & 3.56 & 1.78 & 1.12 & 0.77 & 6.71 & 1.41 & 6.14 & 84.3 \\
& InternVL3.5(8B) & 26.8 & 8.99 & 3.85 & 2.07 & 1.31 & 13.7 & 3.32 & 11.0 & 84.5 \\
& Qwen2.5-VL(7B) & 20.7 & 8.41 & 3.44 & 1.82 & 1.12 & 12.7 & 2.74 & 10.2 & 84.0 \\
& Qwen2.5-VL(32B) & 22.7 & 8.15 & 3.09 & 1.43 & 0.90 & 12.7 & 2.90 & 9.74 & 83.7 \\
& Gemma3(27B) & 21.6 & 8.62 & 3.13 & 1.53 & 0.92 & 12.6 & 2.34 & 10.0 & 84.8 \\
& InternVL2-DA-DriveLM(4B) & 26.4 & 12.2 & 5.29 & 2.80 & 1.82 & 17.4 & 4.15 & 13.7 & 85.4 \\
& Qwen2.5-VL(7B) (FT) & \textbf{44.8} & \textbf{20.0} & \textbf{13.7} & \textbf{7.30} & \textbf{4.92} & \textbf{29.1} & \textbf{15.1} & \textbf{27.3} & \textbf{87.8} \\
\midrule
\multirow{7}{*}{Safety-Critical}
& LLaVA-OneVision(7B) & 25.5 & 5.81 & 2.40 & 1.23 & 0.86 & 11.7 & 2.81 & 10.5 & 85.2 \\
& InternVL3.5(8B) & 20.0 & 9.93 & 4.25 & 2.19 & 1.32 & 14.3 & 3.85 & 11.4 & 84.8 \\
& Qwen2.5-VL(7B) & 23.0 & 9.96 & 3.85 & 1.94 & 1.12 & 14.4 & 3.52 & 10.2 & 84.7 \\
& Qwen2.5-VL(32B) & 19.9 & 6.92 & 2.97 & 1.54 & 0.09 & 11.2 & 2.69 & 8.8 & 82.7 \\
& Gemma3(27B) & 21.3 & 10.8 & 4.38 & 2.18 & 1.32 & 14.9 & 3.53 & 11.7 & 85.1 \\
& InternVL2-DA-DriveLM(4B) & 17.5 & 11.4 & 6.66 & 3.89 & 2.70 & 17.5 & 5.08 & 15.8 & 85.8 \\
& Qwen2.5-VL(7B) (FT) & \textbf{34.0} & \textbf{18.6} & \textbf{9.68} & \textbf{6.56} & \textbf{4.72} & \textbf{26.1} & \textbf{8.20} & \textbf{23.1} & \textbf{87.1} \\
\midrule
\multirow{7}{*}{Overall}
& LLaVA-OneVision(7B) & 25.2 & 6.56 & 3.17 & 1.99 & 1.43 & 11.7 & 3.07 & 10.6 & 85.1 \\
& InternVL3.5(8B) & 25.3 & 10.2 & 4.62 & 2.56 & 1.59 & 14.3 & 3.99 & 11.5 & 84.6 \\
& Qwen2.5-VL(7B) & 25.5 & 10.1 & 4.30 & 2.41 & 1.60 & 14.1 & 3.74 & 11.3 & 84.7 \\
& Qwen2.5-VL(32B) & 25.1 & 8.15 & 3.09 & 1.43 & 0.85 & 12.7 & 3.52 & 10.2 & 83.7 \\
& Gemma3(27B) & 25.5 & 10.8 & 4.47 & 2.33 & 1.45 & 14.6 & 3.53 & 11.7 & 85.0 \\
& InternVL2-DA-DriveLM(4B) & 23.1 & 12.7 & 6.47 & 4.09 & 2.96 & 18.4 & 6.12 & 15.7 & 85.9 \\
& Qwen2.5-VL(7B) (FT) & \textbf{40.4} & \textbf{22.8} & \textbf{13.8} & \textbf{9.38} & \textbf{7.25} & \textbf{31.6} & \textbf{13.4} & \textbf{29.0} & \textbf{88.1} \\
\bottomrule
\end{tabular}}
\caption{Full open-ended evaluation on WaymoQA. We report GPT-based scoring, BLEU-1/2/3/4, ROUGE-1/2/L, and BERTScore for Normal Scene, Others, Safety-Critical Scene, and Overall questions.}
\label{tab:supp_open_ended_full}
\end{table*}

\section{Qualitative Examples of Environmental Augmentation}
\label{app:augmentation_examples}

In the main paper, we briefly describe the use of augmentation to increase robustness to diverse environmental conditions. 
Here, we provide qualitative examples to illustrate the types of appearance changes used in our augmentation pipeline. 
Starting from the same base driving scene, we generate variants that simulate different lighting and weather conditions, including dusk, fog, night, overcast daytime, rain, and snow.

These examples are intended to show the visual diversity introduced by augmentation rather than to define a separate benchmark. 
The goal is to expose the model to appearance shifts that commonly occur in real-world driving, especially those that can affect perception and safety-critical reasoning. 
By augmenting scenes with variations in illumination, visibility, and weather, we improve robustness to domain shifts while preserving the underlying scene layout and driving semantics.

\begin{figure*}[!t]
\centering
\setlength{\tabcolsep}{2pt}
\renewcommand{\arraystretch}{0.85}

\begin{tabular}{cccc}
\includegraphics[width=0.235\textwidth]{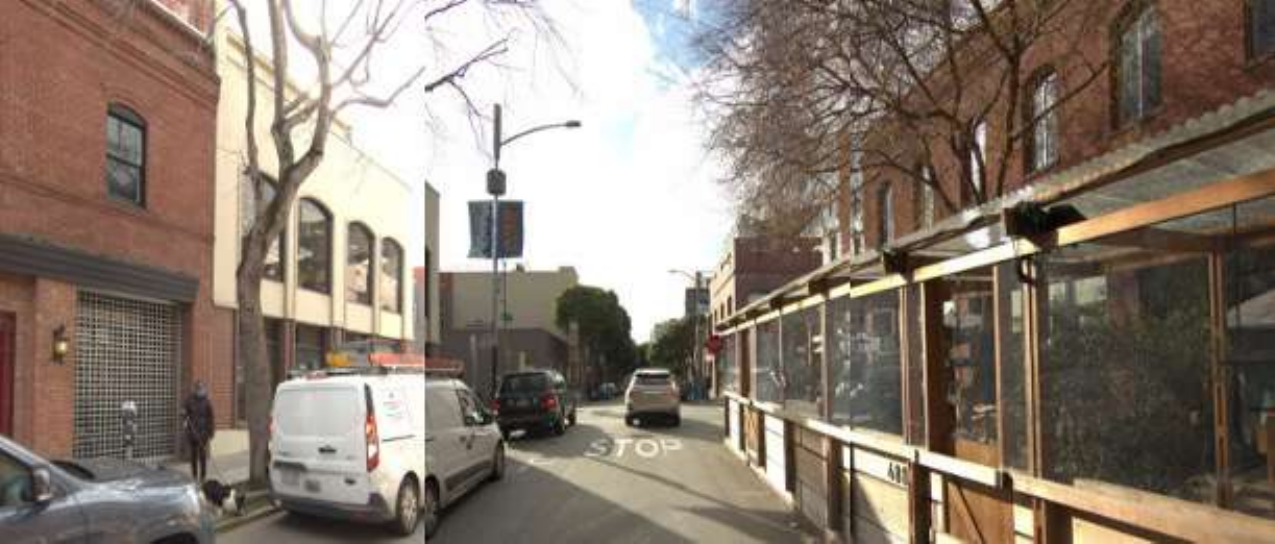} &
\includegraphics[width=0.235\textwidth]{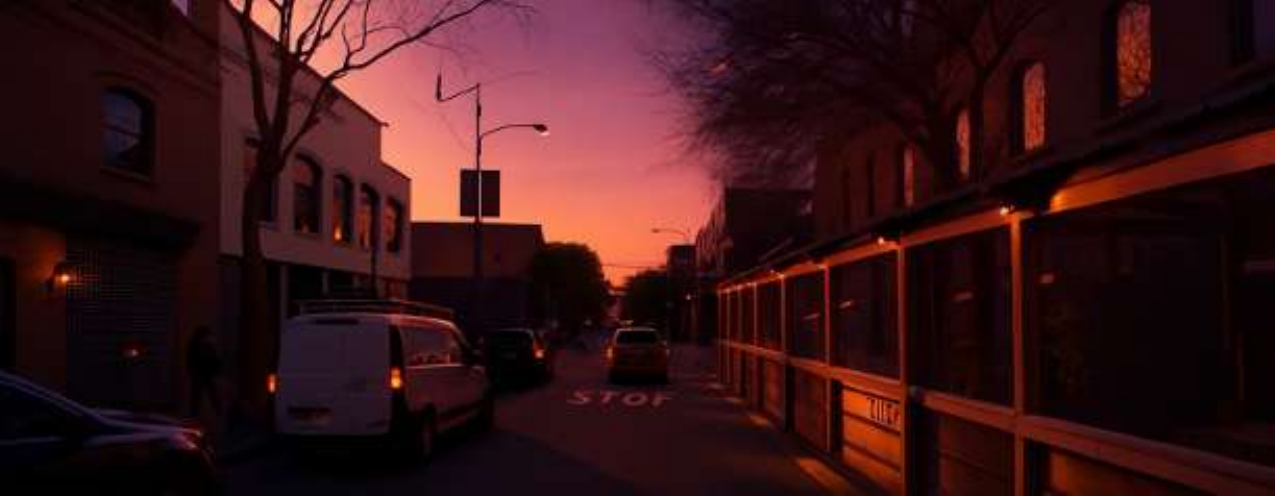} &
\includegraphics[width=0.235\textwidth]{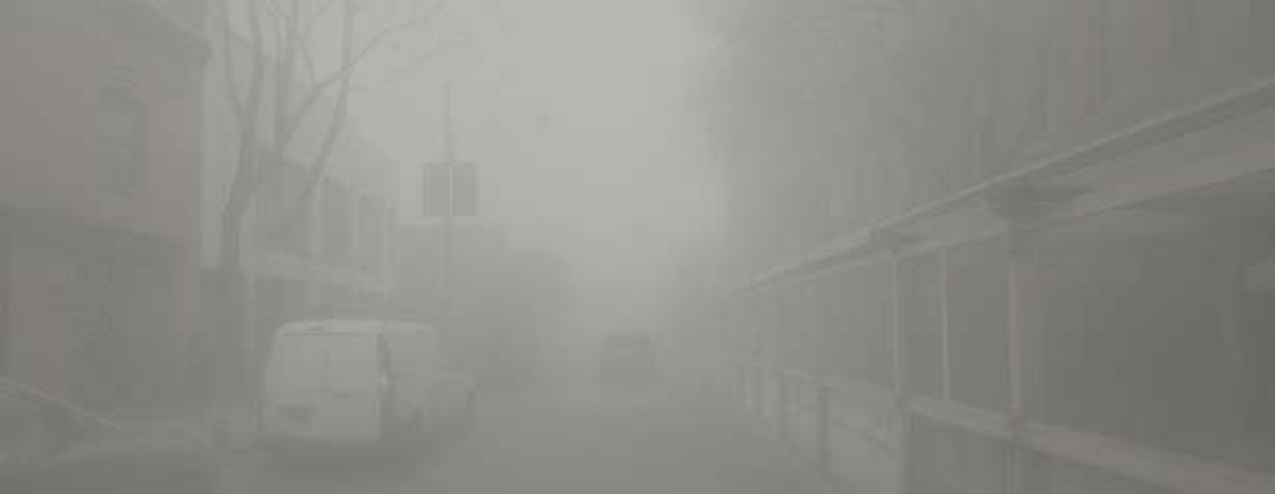} &
\includegraphics[width=0.235\textwidth]{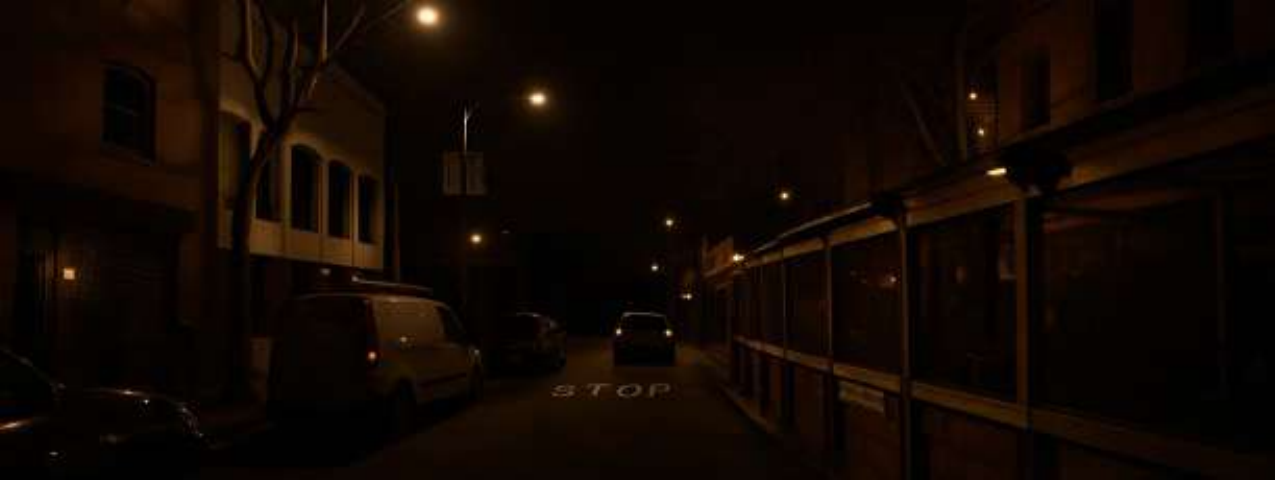} \\
\scriptsize Original & \scriptsize Dusk/Dawn & \scriptsize Foggy & \scriptsize Nighttime \\[2mm]

\includegraphics[width=0.235\textwidth]{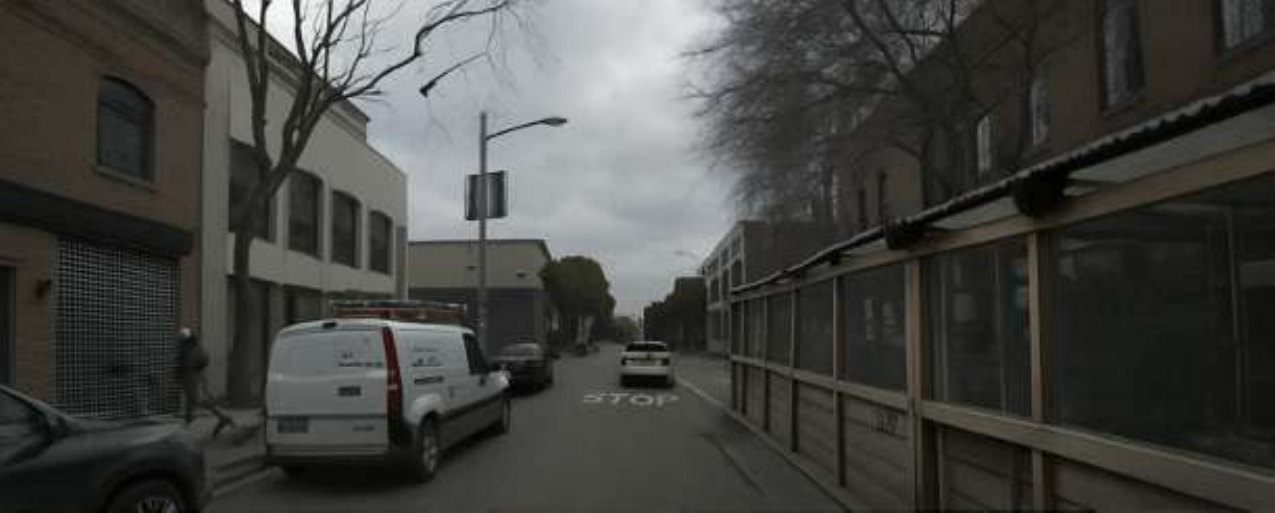} &
\includegraphics[width=0.235\textwidth]{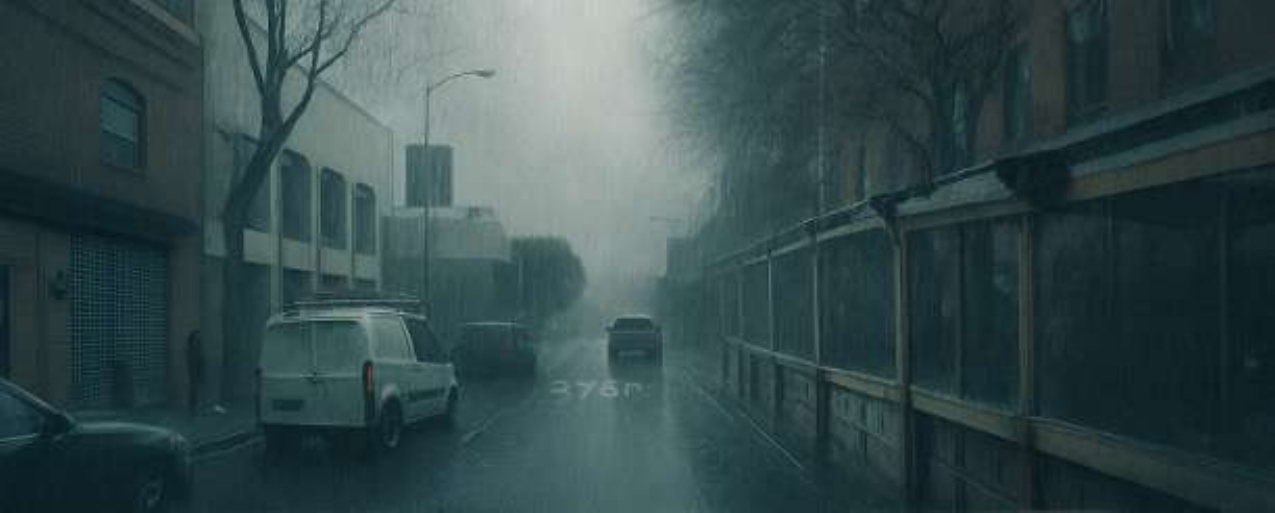} &
\includegraphics[width=0.235\textwidth]{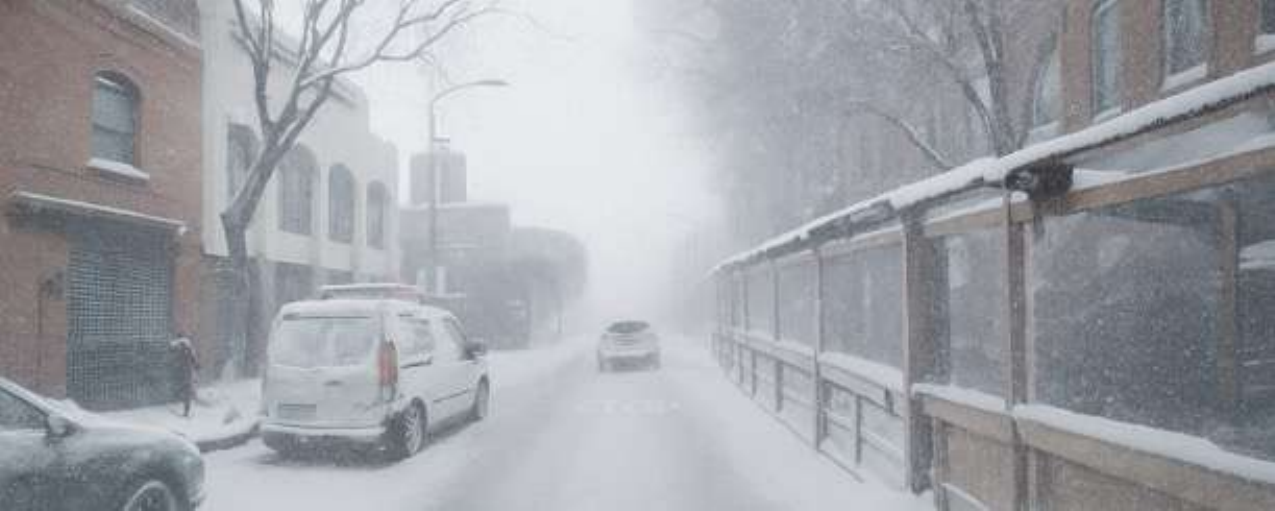} &
\\
\scriptsize Overcast & \scriptsize Rainy & \scriptsize Snowy &
\end{tabular}

\vspace{-2mm}
\caption{Qualitative examples of environmental augmentation applied to the same driving scene. 
The scene structure is preserved while lighting, visibility, and weather conditions change across dusk/dawn, foggy, nighttime, overcast, rainy, and snowy variants.}
\label{fig:augmentation_examples}
\vspace{-3mm}
\end{figure*}

Figure~\ref{fig:augmentation_examples} shows representative augmentation examples generated from a single scene. 
Although the road geometry, object placement, and traffic context remain unchanged, the visual appearance changes substantially across conditions. 
This helps the model learn to rely less on superficial appearance cues and better generalize across diverse environments.

\section{Failure Case.}
\label{sup:fail}
Figure~\ref{fig:failure_rel_frame} illustrates a typical failure pattern of our model.
In the first stage of Safety-Critical Reasoning, the model correctly identifies the primary
safety-critical object (a person and a dog next to the parked vehicle), predicts that the
person will remain standing near the car, and concludes that the ego vehicle should wait
until the person either walks away or re-enters the vehicle.  Thus, basic perception,
prediction, and planning around the first hazard are handled reliably.

However, the second stage exposes a systematic weakness when directions are defined in a
\emph{relative} coordinate frame.  The additional safety-critical object is an oncoming
blue vehicle, described in the ground truth as ``driving in the opposite lane to the left
of the white vehicle in front'' and ``attempting to drive straight ahead to the left of
the white vehicle in front.’’  Our model instead predicts that the blue vehicle is on the
right side of the white vehicle and moving to the right.  In other words, it fails to
reinterpret ``left’’ and ``right’’ from the other vehicle’s perspective and mistakenly
projects them into the ego-centric frame.  This misinterpretation then propagates to
downstream sub-questions: the model’s subsequent plan for “addressing the first risk
while also addressing the second risk’’ is inconsistent with the true geometry of the
scene, leading to an unsafe recommendation.  A similar confusion appears in the normal
reasoning question at the bottom of the figure, where the model again flips the turning
direction of the rear vehicle when the description is given from that vehicle’s viewpoint.

\section{Extending WaymoQA}
\label{sup:extending}
\begin{figure*}
  \centering
  \includegraphics[width=1.0\linewidth]{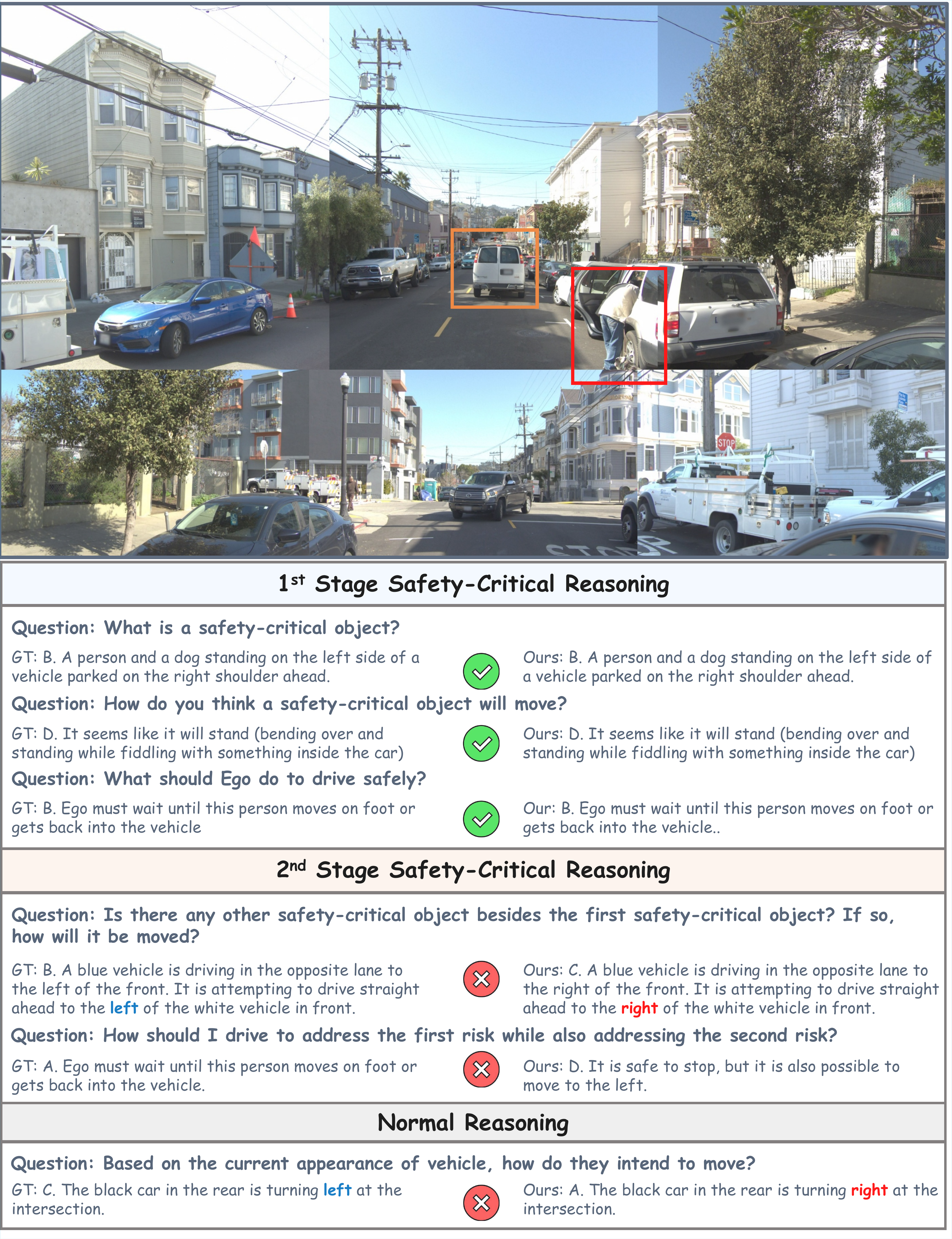}
  \caption{\textbf{Failure case on two–stage Safety-Critical Reasoning in WaymoQA.}}
  \label{fig:failure_rel_frame}
\end{figure*}

Our current benchmark evaluates \emph{perception}, \emph{prediction}, and \emph{planning} with separate questions. 
For example, in the scenario shown in Fig.~\ref{fig:failure_rel_frame}, we ask (i) which object is safety-critical, (ii) how that object will move, and (iii) how the ego vehicle should drive safely. 
While this decomposition reveals which sub-skill fails, it does not explicitly enforce that the model maintains a coherent reasoning chain from perception to planning. 
A model can sometimes answer each sub-question in isolation by pattern matching, without truly understanding how its perception of agents constrains future motion and, in turn, ego decisions.

Because WaymoQA contains rich, temporally aligned annotations, we can instead reorganize these items into \emph{compound} questions that evaluate the full decision process. 
For instance, rather than three separate questions, a single multi-part query could ask:
(1) which agents are safety-critical in the current scene, 
(2) how each of them is likely to move over the next few seconds, and 
(3) which maneuver the ego vehicle should take to reduce overall risk. 
The ground truth already contains all of this information, so the model’s answer can be graded both holistically and at the level of partial credit: a model that correctly identifies the primary hazard but mispredicts the second agent’s motion would receive credit for perception but be penalized on prediction and planning. 
Such structured supervision directly ties mistakes in downstream planning to specific upstream errors, e.g., misinterpreting a relative left/right description for an oncoming vehicle.

We view these extensions as a natural next step for WaymoQA. 
By transforming our existing annotations into multi-stage, chain-structured questions with explicit partial scoring, future work can train and test models on \emph{complete} safety-critical reasoning pipelines rather than isolated sub-tasks, and thus more directly target the types of compounding failures observed in our error analysis.

\end{document}